\documentclass[11pt,a4paper]{style/arxiv-template}

\usepackage{flafter,adjustbox,enumitem,fvextra,needspace}

\newcommand{\bpInlineCode}[1]{{\fontencoding{T1}\ttfamily\selectfont #1}}
\newcommand{\bpBacktick}{{\fontencoding{T1}\selectfont\char0}}
\newcommand{\bpSingleQuote}{{\fontencoding{TS1}\selectfont\char39}}
\makeatletter
\newcommand{\bpHeading}[1]{\par\addvspace{4pt}\noindent\textbf{#1}\par\nobreak\@afterheading}
\makeatother
\newcommand{\bpGap}{\par\addvspace{3pt}}
\newcommand{\bpTemplateLine}[1]{\noindent\bpInlineCode{#1}\par}
\newenvironment{bpLiteral}{%
  \par\noindent\begin{minipage}{\linewidth}
  \small\fontencoding{T1}\ttfamily\selectfont\raggedright\setlength{\parskip}{0pt}
}{\end{minipage}\par}
\newenvironment{bpList}{\begin{list}{\textbullet}{\setlength{\leftmargin}{12pt}\setlength{\labelsep}{4pt}\setlength{\itemsep}{1pt}\setlength{\parsep}{0pt}\setlength{\topsep}{2pt}}}{\end{list}}

\tcbset{benchmark prompt/.style={
  enhanced,breakable,boxrule=.6pt,arc=2pt,boxsep=0pt,
  colback=white,colframe=black,width=\linewidth,
  enlarge left by=0pt,enlarge right by=0pt,
  left=9pt,right=9pt,top=7pt,bottom=7pt,before skip=6pt,after skip=6pt,
  colbacktitle=black,coltitle=white,
  fonttitle=\small\bfseries,
  toptitle=4pt,bottomtitle=4pt,pad at break*=3pt
}}
\newenvironment{benchmarkprompt}[1]{%
  \begin{tcolorbox}[benchmark prompt,title={#1},
    title after break={#1\hfill\normalfont\itshape continued}]
  \small\setlength{\parskip}{0pt}\raggedright
  \hyphenpenalty=10000\exhyphenpenalty=10000
}{\end{tcolorbox}}
\newenvironment{benchmarktool}[1]{%
  \begin{tcolorbox}[benchmark prompt,title={#1 Bash Tool Definition},breakable=false]
}{\end{tcolorbox}}

\settemplatecolor{templateRose}

\title{RSI-Router: Evolving Subtask-Level LLM Routing and Skills for Cost-Efficient Agents\par}
\author{Hao~Li\footnote[1]{This work was done during their internship at Shanghai Artificial Intelligence Laboratory.}\footnote[3]{\label{fn:nwpu-affiliation}Hao Li and Danyang Jia are affiliated with Northwestern Polytechnical University.}}
\author{Hangfan~Zhang}
\author{Zhiyao~Cui}
\author{Chunjiang~Mu}
\author{\\Yiqun~Zhang}
\author{Bo~Zhang}
\author{Danyang~Jia\textsuperscript{\ref{fn:nwpu-affiliation}}}
\author{Shuyue~Hu\footnote[2]{Corresponding author.}}
\affiliation{Shanghai Artificial Intelligence Laboratory}
\contribution{\email{lihao4@pjlab.org.cn}\quad\email{hushuyue@pjlab.org.cn}}

\abstract{Practical deployment of large language model (LLM) agents requires strong task performance at affordable inference cost.
For long-horizon agentic tasks, this performance--cost trade-off can be improved through within-task large--small model collaboration, as smaller models can handle some stages even when they cannot solve the full task.
In this paper, we introduce \textit{RSI-router}, a routing framework that constructs subtask-level model assignments and model-specific skills through recursive self-improvement over accumulated experience.
Each iteration consists of four stages: \textit{Subtask Mining} derives subtask definitions and identification rules from training trajectories; \textit{Routing Strategy Evolution} proposes and evaluates diverse model assignments; \textit{Model-Specific Skill Evolution} compares routed and large-model-only trajectories to diagnose failures and develop reusable execution skills; and \textit{Pareto-Optimal Router Selection} updates the Pareto population using historical and newly generated routers while retaining dominated routers as experience for subsequent evolution.
\textbf{Routing between DeepSeek-V4.1-Flash and Qwen3.5-9B, RSI-router consistently surpasses the DeepSeek-only baseline at roughly half the inference cost (48.3\%) across five agentic benchmarks}. In particular, on ALFWorld, ScienceWorld, and WebShop, it cuts inference cost by \textbf{74.7--82.2\%} while simultaneously improving performance; on Terminal-Bench 2.0, it achieves a \textbf{16.7\%} relative performance gain at \textbf{18.0\%} lower cost.
Moreover, RSI-router establishes a stronger performance--cost Pareto frontier than 9 routing methods.}
\keywords{LLM routing, Cost-efficient inference, Agentic tasks, Recursive self-improvement}

\hypersetup{
  pdftitle={RSI-Router: Cost-Efficient LLM Routing via Evolution of Subtask-Level Model Assignments with Execution Skills},
  pdfauthor={Hao Li, Hangfan Zhang, Zhiyao Cui, Chunjiang Mu, Yiqun Zhang, Bo Zhang, Danyang Jia, Shuyue Hu},
  pdfsubject={Cost-efficient LLM routing through subtask-level model assignments and execution skills}
}

\begin{document}
\begingroup
\renewcommand{\thefootnote}{\fnsymbol{footnote}}
\maketitle
\endgroup

\section{Introduction}
\label{sec:intro}

\begin{figure}[t]
\centering
\includegraphics[width=\linewidth]{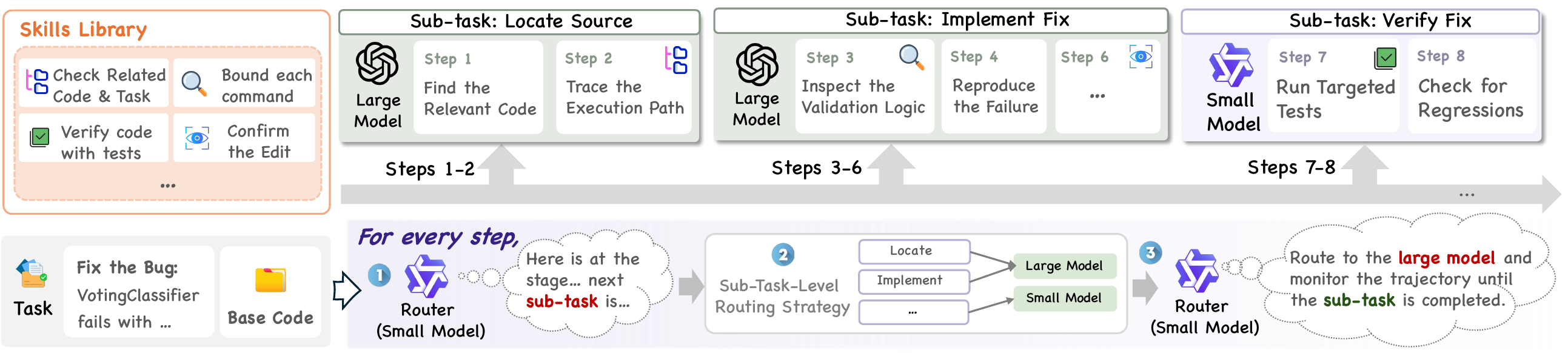}
\caption{\textbf{Illustration of RSI-router execution.}
At each step, a small model serves as the router, identifying the upcoming subtask from the routing context. The router selects a model for the subtask and appends the corresponding model-specific skills to the selected model’s input.
In this SWE-bench example, the large model locates relevant code and implements the fix, while the small model verifies it.}
\label{fig:rsi-infer}
\end{figure}

Practical deployment of large language model (LLM) agents requires a balance between strong task performance and cost-efficient inference, especially for complex agentic tasks such as software development~\citep{liu2024large,yan2026harness}, personal assistance~\citep{sager2026comprehensive,cui2026design}, and scientific research~\citep{lu2026towards,cui2026agentpanel}.
These tasks often unfold over long interaction sequences and require repeated LLM calls, making exclusive reliance on advanced, larger models costly despite their strong capabilities~\citep{gao2026more}.
As smaller LLMs become increasingly capable~\citep{qwen2026qwen35,abouelenin2025phi}, large--small model collaboration offers a promising way to improve this performance--cost trade-off by delegating suitable parts of a task to smaller models while reserving larger models for parts that require stronger capabilities~\citep{alazraki2026scaling}.

LLM routing~\citep{hu2024routerbench,li2026llmrouterbench} enables such collaboration by adaptively selecting which model to use.
Task-level methods~\citep{jiang2023llm,ding2024hybrid,jitkrittum2026universal} typically assign an entire request to a single model.
However, agentic tasks often contain stages that smaller models can handle effectively, even when they cannot complete the full task.
Assigning one model throughout therefore misses opportunities to reduce cost through within-task collaboration.
Step-level methods address this limitation by selecting a model at each interaction step, but learning effective routing decisions remains challenging.
Some methods~\citep{zhang2026mtrouter,wang2026skillorchestra} learn from pre-collected trajectories and are constrained by the coverage of those trajectories.
Others explore model assignment through reinforcement learning~\citep{zhang2025routerr,zhang2026budget}, but can face costly stepwise exploration without task-structure guidance and difficult credit assignment from final-outcome feedback~\citep{zhang2024advancing,tan2026papo,tan2026scaling}.

To address these challenges, we introduce \textit{RSI-router}, a routing framework that recursively improves its routing strategies and execution skills using execution outcomes and accumulated experience.
RSI-router is motivated by three observations.
First, LLMs typically possess prior knowledge of task objectives and procedures, which can organize the exploration of model assignments around reusable subtasks rather than treating every interaction step as an independent decision.
Second, execution trajectories reveal where a particular model fails, succeeds, or performs redundant actions, providing concrete evidence for revising model assignments and developing model-specific execution guidance.
Third, such guidance can itself change what a model can reliably accomplish: a smaller model that initially fails on a subtask may become effective when equipped with an appropriate execution skill.
Consequently, model assignments and execution skills are interdependent and need to be improved jointly rather than optimized in isolation.




Specifically, RSI-router recursively improves the routing system through a four-stage evolution loop.
\textbf{Subtask Mining} derives and revises subtask identification from trajectories.
\textbf{Routing Strategy Evolution} proposes diverse routing strategies over these subtasks in parallel and evaluates candidate routers with selectively inherited skills.
\textbf{Model-Specific Skill Evolution} compares candidate and large-model-only trajectories on the same tasks to diagnose model-specific failure patterns and redundant actions, then refines or generates execution skills tailored to subtask--model pairs.
\textbf{Pareto-Optimal Router Selection} compares historical and newly generated routers by performance and cost, updating the Pareto population while retaining dominated routers as negative examples for subsequent evolution.
The resulting routers, skills, and execution experience further guide subsequent iterations.
While a large LLM drives this evolution process, at inference time, a small model serves as the router, identifying the upcoming subtask, selecting its assigned model, and appending the corresponding execution skills to the selected model's input (Figure~\ref{fig:rsi-infer}).


We evaluate RSI-router on five benchmarks of agentic tasks:
ALFWorld~\citep{shridhar2021alfworld},
ScienceWorld~\citep{wang2022scienceworld},
WebShop~\citep{yao2022webshop},
SWE-bench Verified~\citep{jimenez2024swe,chowdhury2024swebenchverified},
and Terminal-Bench 2.0~\citep{merrill2026terminal}.
Across these benchmarks, RSI-router reduces inference cost by \textbf{8.1--82.2\%},
with an average reduction of \textbf{51.7\%}, while matching or exceeding
large-model-only performance.
For example, it improves ALFWorld success rate from 90.62\% to 98.44\%
with 75.3\% lower cost, and Terminal-Bench 2.0 from 40.00\% to 46.67\%
with 18.0\% lower cost.
RSI-router also improves the performance--cost Pareto frontier over the
evaluated task-level and step-level routing baselines.
Further analysis shows repeated cost reductions over iterations
and continued reuse of skills acquired in earlier iterations, while ablation studies further corroborate key routing and execution design choices.

\begin{figure}[t]
\centering
\includegraphics[width=\linewidth]{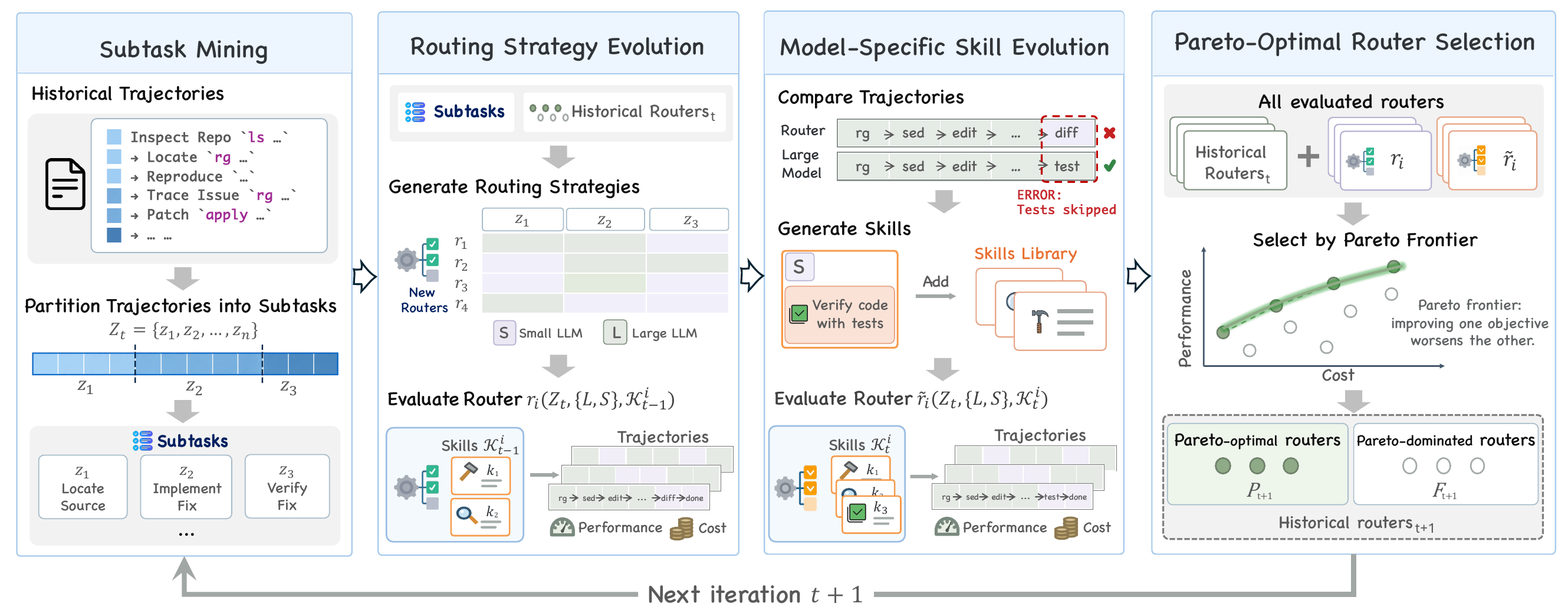}
\caption{\textbf{Router evolution in RSI-router.}
A proposal agent iteratively mines subtasks, explores model assignments with selectively inherited skills, and evolves model-specific skills by comparing training trajectories with large-model-only executions.
Pareto-optimal and dominated routers are both retained to guide subsequent iterations.}
\vspace{-1em}
\label{fig:rsi-framework}
\end{figure}

Our main contributions are as follows:

\begin{itemize}[leftmargin=1.2em]
\item \textbf{A recursive self-improvement paradigm for LLM routing.}
We introduce \textit{RSI-router}, which constructs cost-efficient routing strategies through iterative improvement over execution outcomes and accumulated experience.

\item \textbf{Subtask-level routing with model-specific execution skills.}
RSI-router discovers reusable subtasks, evolves model assignments at the subtask level, and develops model-specific execution skills, enabling smaller models to reliably handle suitable stages of long-horizon agentic tasks.

\item \textbf{Advancing the Pareto frontier across five benchmarks.}
Across five agentic benchmarks, RSI-router matches or exceeds the performance of the large-model baseline while reducing inference cost by 51.7\% on average, and consistently extends the performance--cost Pareto frontier beyond existing routing baselines.
\end{itemize}

\section{Method}
\label{sec:method}


Agentic tasks comprise stages with different capability requirements, and the most cost-effective model may vary across stages.
We group execution steps with similar objectives into subtasks to explore model assignments and reuse experience.
Evaluating these assignments yields trajectories revealing model-specific failures and redundant actions, guiding the development of execution skills.
As these skills change models' effective capabilities, model assignments need to be revisited.
We therefore propose \textit{RSI-router}, a routing framework that refines subtask definitions, model assignments, and execution guidance through iterative evolution to improve performance--cost trade-offs.

Starting with training trajectories from large-model-only and small-model-only executions, RSI-router evolves a population of routers through a four-stage loop guided by a large-model proposal agent (Figure~\ref{fig:rsi-framework}):
\textbf{Subtask Mining} organizes execution experience into reusable subtasks to guide model assignments;
\textbf{Routing Strategy Evolution} explores subtask-level model assignments to generate candidate routers with different performance--cost trade-offs;
\textbf{Model-Specific Skill Evolution} distills the resulting training trajectories into reusable execution skills to extend models' effective capabilities; and
\textbf{Pareto-Optimal Router Selection} guards against regressions by comparing historical and new routers using validation performance and cost.




\subsection{Routing and Evolution Setup}
\label{sec:routing-setup}


We consider agentic task execution using a large model $L$ and a small model $S$, supported by reusable execution skills.
To explore how the two models should collaborate, we evolve a population of routers with different model assignments and execution skills.
At iteration $t$, newly proposed candidates share a subtask set $Z_t$, with definitions describing execution objectives and identification rules for predicting the upcoming subtask.
Each candidate router $r_i$ specifies which model executes each subtask; for example, $L$ may implement a fix while $S$ verifies it.
Each skill contains execution instructions and specifies the applicable subtasks and execution model.

Before each execution step, the small model $S$ predicts the upcoming step's subtask using the definitions and identification rules in $Z_t$ and the routing context $c$.
This context contains the user query, the interaction history from the previous few steps, and the previous subtask prediction.
The router then selects the assigned model, appends only the skill text matching the predicted subtask and selected model to its input, and invokes it to generate the response.


\textbf{Router evolution overview.}
We run router evolution for $G$ iterations to maximize task performance and minimize inference cost.
We maintain a Pareto population $\mathcal P_t$ of non-dominated routers and an archive $\mathcal F_t$ of dominated routers retained as negative examples for subsequent refinement, initialized as $\mathcal P_0=\{r_L,r_S\}$ and $\mathcal F_0=\emptyset$.
The baselines $r_L$ and $r_S$ use $L$ and $S$, respectively, throughout execution without skills.
Skills from historical routers in $\mathcal P_t \cup \mathcal F_t$ form a shared library for reuse.
Each router $r$ is evaluated on disjoint training and validation sets to assess whether training-derived refinements remain effective on other tasks.
Aggregate performance and cost from both sets guide evolution, while execution trajectories $\mathcal T(r)$ and per-task feedback are provided only for training tasks.
Validation performance and cost guide Pareto-based population updates, yielding the final population $\mathcal P_G$.

\subsection{Router Evolution}
\label{sec:self-evolution}

\paragraph{Subtask Mining.}
To construct $Z_t$, we identify subtasks in training trajectories and derive their definitions and identification rules.
For example, code repair may involve locating relevant code, implementing a fix, and verifying the result, with the same subtask recurring within a trajectory.

In iteration $t$, the proposal agent first analyzes these trajectories and partitions them into segments by execution objective.
It then defines and clusters subtasks across trajectories to form $Z_t$, assigning subtask labels to segments and deriving identification rules.
From the second iteration onward, the agent revisits previously defined subtasks, retaining, splitting, or merging them and updating the corresponding definitions and identification rules.

To check whether the resulting subtasks can be identified before execution, we replay sampled training trajectories.
At each step, the small model $S$ predicts the subtask label from the pre-execution context $c$ using the current definitions and identification rules.
The proposal agent compares these predictions with the annotated labels and refines ambiguous definitions or rules.
The finalized $Z_t$ remains fixed for the rest of the iteration.

\paragraph{Routing Strategy Evolution.}
With $Z_t$ established, we explore model assignments with different performance--cost trade-offs.
In each iteration, the proposal agent generates $N$ diverse routing strategies in parallel, indexed by $i=1,\ldots,N$.
It draws on the model assignments, associated subtask definitions, and aggregate training and validation performance and cost of historical routers to explore alternative assignments.

For the $i$-th strategy, the proposal agent selectively inherits skills from the shared library based on their relevance to the current subtask definitions and model assignments.
It reassesses applicability after subtask splits or merges, selecting zero or more skills per subtask to form $\mathcal K_{t-1}^i$.
Each candidate router is evaluated on the training and validation sets to measure performance and cost.
Its training trajectories and per-task feedback are passed to the next stage to guide model-specific skill evolution.



\paragraph{Model-Specific Skill Evolution.}
Execution skills complement model assignments by providing reusable, model-specific guidance to help models complete their assigned subtasks reliably and avoid redundant actions.
As subtask definitions and model assignments evolve, newly collected trajectories reveal where this guidance needs further refinement.
For each candidate router $r_i$, the proposal agent compares its training trajectories with those of $r_L$ on the same tasks, using per-task feedback and cost to identify model-specific failure patterns and redundant actions.

Based on this diagnosis, the agent refines inherited skills or generates new ones for the relevant subtasks and assigned models, updating $\mathcal K_{t-1}^{i}$ to $\mathcal K_t^{i}$.
We denote the updated router by $\tilde r_i$ and re-evaluate it on the training and validation sets, keeping $Z_t$ and the model assignments unchanged.


\paragraph{Pareto-Optimal Router Selection.}
To guide evolution towards higher task performance and lower inference cost while guarding against regressions, we jointly compare historical routers in $\mathcal P_t \cup \mathcal F_t$ and candidates before and after skill evolution using validation performance and cost.
The non-dominated routers form $\mathcal P_{t+1}$, while the remaining routers are archived in $\mathcal F_{t+1}$.
Routers in both sets remain available for proposing routing strategies and inheriting execution skills in subsequent iterations.
After $G$ iterations, RSI-router returns the final Pareto population $\mathcal P_G$.

\section{Experiments}
\label{sec:experiments}

We evaluate RSI-router on five benchmarks of agentic tasks to examine whether it can reduce online cost without sacrificing task performance. We further analyze the router evolution process and assess key design choices through ablation studies.

\subsection{Setup}
\label{sec:experimental-setup}

\paragraph{Compared methods.}
We compare against the large-model-only router $r_L$, the small-model-only router $r_S$, and random routers that select either model with equal probability at the task or step level.
The task-level routing methods include HybridLLM~\citep{ding2024hybrid}, FrugalGPT~\citep{chen2024frugalgpt}, RouteLLM~\citep{ong2025routellm}, GraphRouter~\citep{feng2025graphrouter}, and Avengers-Pro~\citep{zhang2025beyond}.
We also include Router-R1~\citep{zhang2025routerr} and MTRouter~\citep{zhang2026mtrouter} as step-level routing baselines (see Appendix~\ref{app:baseline-implementation} for details).

\paragraph{Benchmarks.}
We use ALFWorld~\citep{shridhar2021alfworld}, ScienceWorld~\citep{wang2022scienceworld}, WebShop~\citep{yao2022webshop}, SWE-bench Verified~\citep{jimenez2024swe,chowdhury2024swebenchverified}, and Terminal-Bench 2.0~\citep{merrill2026terminal}, covering household environment interaction, scientific experimentation, online shopping, code repository repair, and terminal operations, respectively. The full benchmark prompts and tool definitions are provided
in Appendix~\ref{app:benchmark-prompts}.

\paragraph{Models and runtime settings.}
Across all five benchmarks, we use DeepSeek-V4.1-Flash~\citep{deepseek2026v41flash} as the large model $L$ and Qwen3.5-9B~\citep{qwen2026qwen35} as the small model $S$.
Our execution harness is built on mini-SWE-agent~\citep{yang2024swe}, with environment interactions adapted to each benchmark.
We use Codex~\citep{openai2025codex} as the proposal agent and prompt $S$ for subtask identification using the interaction history from the previous 3 steps.

\paragraph{Data splits and iteration protocol.}
We use fixed, disjoint training, validation, and test sets ($\mathcal D_{\mathrm{tr}}$, $\mathcal D_{\mathrm{val}}$, $\mathcal D_{\mathrm{test}}$), with 64 tasks per set for ALFWorld, ScienceWorld, and WebShop, 32 per set for SWE-bench Verified, and 32/32/25 tasks for Terminal-Bench 2.0.
Router evolution uses aggregate performance and cost from training and validation, but trajectories and per-task feedback only from training.
Validation metrics guide candidate router generation and Pareto population updates.
After router evolution, we evaluate all routers in the final Pareto population $\mathcal P_G$ on $\mathcal D_{\mathrm{test}}$, which provides no feedback for router evolution.
We run router evolution for $G=6$ iterations with $N=4$ candidate routing strategies per iteration.

\paragraph{Evaluation metrics.}
We report task success rate (\%) on all benchmarks except WebShop, where we use reward (0--1). Costs are reported in USD using the official DeepSeek-V4.1-Flash input and output prices. We estimate the corresponding small-model prices using a fixed $1{:}30$ reference ratio informed by GPU-hour measurements on H200 GPUs (see Appendix~\ref{app:gpu-hour-cost} for details). Costs include all test-time LLM calls and tokens, including subtask routing and skill context. All results are averaged over three evaluation repeats per test task. For learned baselines, we additionally average over three training seeds.

\begin{table}[htbp]
\begingroup
\centering
\setlength{\abovecaptionskip}{2pt}
\setlength{\belowcaptionskip}{5pt}


\caption{
\textbf{RSI-router matches or exceeds the large-model baseline at substantially lower inference cost.} Cost is total test-set inference cost (USD). $\Delta$ denotes relative percentage changes from the indicated baseline.
}
\label{tab:rsi-9b-grouped}

\definecolor{oursTint}{HTML}{EBE8FA}
\definecolor{comparisonTint}{HTML}{F6F5FC}
\definecolor{performanceBlue}{HTML}{243E91}
\definecolor{costOrange}{HTML}{A85408}
\definecolor{negativeRed}{HTML}{B12C2C}
\definecolor{upGreen}{HTML}{176F0B}
\definecolor{groupInk}{HTML}{000000}

\fontsize{9}{11}\selectfont
\setlength{\tabcolsep}{2.5pt}
\renewcommand{\arraystretch}{1.12}

\newcommand{\rsiTableHead}[1]{%
  \raisebox{\dimexpr(\depth-\height)/2\relax}[12pt][12pt]{%
    \bfseries\shortstack[c]{#1}%
  }%
}

\begin{adjustbox}{max width=\linewidth}
\begin{tabular}{l*{5}{rr}}
\toprule

\multicolumn{1}{c}{\rsiTableHead{Method}}
& \multicolumn{2}{c}{\rsiTableHead{ALFWorld}}
& \multicolumn{2}{c}{\rsiTableHead{ScienceWorld}}
& \multicolumn{2}{c}{\rsiTableHead{WebShop}}
& \multicolumn{2}{c}{\rsiTableHead{SWE-bench\\Verified}}
& \multicolumn{2}{c}{\rsiTableHead{Terminal-Bench\\2.0}} \\

\cmidrule(lr){2-3}
\cmidrule(lr){4-5}
\cmidrule(lr){6-7}
\cmidrule(lr){8-9}
\cmidrule(lr){10-11}

& \multicolumn{1}{c}{\fontsize{8.5}{10}\selectfont Perf.\,{\color{upGreen}$\uparrow$}}
& \multicolumn{1}{c}{\fontsize{8.5}{10}\selectfont Cost\,{\color{negativeRed}$\downarrow$}}
& \multicolumn{1}{c}{\fontsize{8.5}{10}\selectfont Perf.\,{\color{upGreen}$\uparrow$}}
& \multicolumn{1}{c}{\fontsize{8.5}{10}\selectfont Cost\,{\color{negativeRed}$\downarrow$}}
& \multicolumn{1}{c}{\fontsize{8.5}{10}\selectfont Perf.\,{\color{upGreen}$\uparrow$}}
& \multicolumn{1}{c}{\fontsize{8.5}{10}\selectfont Cost\,{\color{negativeRed}$\downarrow$}}
& \multicolumn{1}{c}{\fontsize{8.5}{10}\selectfont Perf.\,{\color{upGreen}$\uparrow$}}
& \multicolumn{1}{c}{\fontsize{8.5}{10}\selectfont Cost\,{\color{negativeRed}$\downarrow$}}
& \multicolumn{1}{c}{\fontsize{8.5}{10}\selectfont Perf.\,{\color{upGreen}$\uparrow$}}
& \multicolumn{1}{c}{\fontsize{8.5}{10}\selectfont Cost\,{\color{negativeRed}$\downarrow$}} \\

\midrule
\rowcolor[HTML]{FBF1F1}
\multicolumn{11}{c}{%
  \rule{0pt}{2.5ex}\fontsize{8.5}{10}\selectfont
  \color{groupInk}\bfseries\itshape Single-model baselines
} \\

DeepSeek-V4.1-Flash
& 90.62 & 1.50
& 34.38 & 1.91
& 0.59 & 0.75
& 83.33 & 3.46
& 40.00 & 17.10 \\

Qwen3.5-9B
& 65.62 & 0.13
& 14.58 & 0.06
& 0.48 & 0.02
& 62.50 & 0.69
& 5.33 & 0.17 \\

\midrule
\rowcolor[HTML]{F2F1FD}
\multicolumn{11}{c}{%
  \rule{0pt}{2.5ex}\fontsize{8.5}{10}\selectfont
  \color{groupInk}\bfseries\itshape Task-level routing
} \\

Random & 79.69 & 0.82 & 26.56 & 0.72 & 0.55 & 0.43 & 64.58 & 1.23 & 26.67 & 8.22 \\

HybridLLM
& 68.92 & 0.46
& 25.69 & 0.44
& 0.51 & 0.14
& 69.79 & 2.33
& 24.00 & 8.62 \\

FrugalGPT
& 86.81 & 1.21
& 31.42 & 1.53
& 0.57 & 0.69
& 68.40 & 2.07
& 21.78 & 9.52 \\

RouteLLM
& 84.38 & 1.16
& 32.99 & 1.55
& 0.59 & 0.67
& 72.92 & 2.68
& 40.00 & 17.10 \\

GraphRouter
& 90.62 & 1.50
& 34.38 & 1.91
& 0.59 & 0.75
& 83.33 & 3.46
& 40.00 & 17.10 \\

Avengers-Pro
& 90.62 & 1.50
& 33.33 & 1.71
& 0.59 & 0.75
& 79.17 & 3.33
& 40.00 & 17.10 \\

\midrule
\rowcolor[HTML]{FEF8E8}
\multicolumn{11}{c}{%
  \rule{0pt}{2.5ex}\fontsize{8.5}{10}\selectfont
  \color{groupInk}\bfseries\itshape Step-level routing
} \\

Random
& 84.38 & 1.24
& 20.83 & 0.72
& 0.53 & 0.34
& 83.33 & 3.89
& 28.00 & 6.87 \\

Router-R1
& 72.92 & 0.19
& 20.31 & 0.11
& 0.49 & 0.23
& 55.21 & 0.28
& 10.67 & 0.31 \\

MTRouter
& 85.94 & 1.06
& 31.08 & 1.00
& 0.54 & 0.41
& 71.88 & 2.38
& 24.89 & 5.85 \\

\addlinespace[2pt]
\rowcolor{oursTint}
\textbf{RSI-router}
& \bfseries 98.44 & \bfseries 0.37
& \bfseries 35.94 & \bfseries 0.34
& \bfseries 0.61 & \bfseries 0.19
& \bfseries 84.38 & \bfseries 3.18
& \bfseries 46.67 & \bfseries 14.02 \\

\rowcolor{comparisonTint}
\hspace{0.4em}{\fontsize{8.5}{10}\selectfont\itshape $\Delta$ vs DeepSeek}
& {\color{performanceBlue}\fontsize{8.5}{10}\selectfont\textbf{+8.6\%}}
& {\color{costOrange}\fontsize{8.5}{10}\selectfont $\mathit{-75.3\%}$}
& {\color{performanceBlue}\fontsize{8.5}{10}\selectfont\textbf{+4.5\%}}
& {\color{costOrange}\fontsize{8.5}{10}\selectfont $\mathit{-82.2\%}$}
& {\color{performanceBlue}\fontsize{8.5}{10}\selectfont\textbf{+3.4\%}}
& {\color{costOrange}\fontsize{8.5}{10}\selectfont $\mathit{-74.7\%}$}
& {\color{performanceBlue}\fontsize{8.5}{10}\selectfont\textbf{+1.3\%}}
& {\color{costOrange}\fontsize{8.5}{10}\selectfont $\mathit{-8.1\%}$}
& {\color{performanceBlue}\fontsize{8.5}{10}\selectfont\textbf{+16.7\%}}
& {\color{costOrange}\fontsize{8.5}{10}\selectfont $\mathit{-18.0\%}$} \\

\rowcolor{comparisonTint}
\hspace{0.4em}{\fontsize{8.5}{10}\selectfont\itshape $\Delta$ vs MTRouter}
& {\color{performanceBlue}\fontsize{8.5}{10}\selectfont\textbf{+14.5\%}}
& {\color{costOrange}\fontsize{8.5}{10}\selectfont $\mathit{-65.1\%}$}
& {\color{performanceBlue}\fontsize{8.5}{10}\selectfont\textbf{+15.6\%}}
& {\color{costOrange}\fontsize{8.5}{10}\selectfont $\mathit{-66.0\%}$}
& {\color{performanceBlue}\fontsize{8.5}{10}\selectfont\textbf{+13.0\%}}
& {\color{costOrange}\fontsize{8.5}{10}\selectfont $\mathit{-53.7\%}$}
& {\color{performanceBlue}\fontsize{8.5}{10}\selectfont\textbf{+17.4\%}}
& {\color{costOrange}\fontsize{8.5}{10}\selectfont $\mathit{+33.6\%}$}
& {\color{performanceBlue}\fontsize{8.5}{10}\selectfont\textbf{+87.5\%}}
& {\color{costOrange}\fontsize{8.5}{10}\selectfont $\mathit{+139.7\%}$} \\

\bottomrule
\end{tabular}
\end{adjustbox}
\endgroup
\end{table}

\subsection{Main Results}
\label{sec:main-results}

\textbf{RSI-router matches or exceeds the large-model baseline at substantially lower inference cost.}
Table~\ref{tab:rsi-9b-grouped} compares RSI-router with single-model, task-level, and step-level baselines across five benchmarks.
Since some methods may produce multiple routing strategies, we report, for each method, the minimum inference cost among strategies that achieve \emph{at least} the performance of the large-model baseline (DeepSeek-V4.1-Flash), ensuring a fair comparison under the same performance requirement.
RSI-router exceeds DeepSeek-V4.1-Flash in performance across all five benchmarks, while reducing inference cost by 8.1--82.2\%.
On ALFWorld, success rate increases from 90.62\% to 98.44\% with a 75.3\% cost reduction.
Across ALFWorld, ScienceWorld, and WebShop, RSI-router surpasses the large model at around one quarter of its cost or less, demonstrating that large--small model collaboration can improve performance and cost efficiency simultaneously.

\begin{figure}[htbp]
\centering
\includegraphics[width=\linewidth]{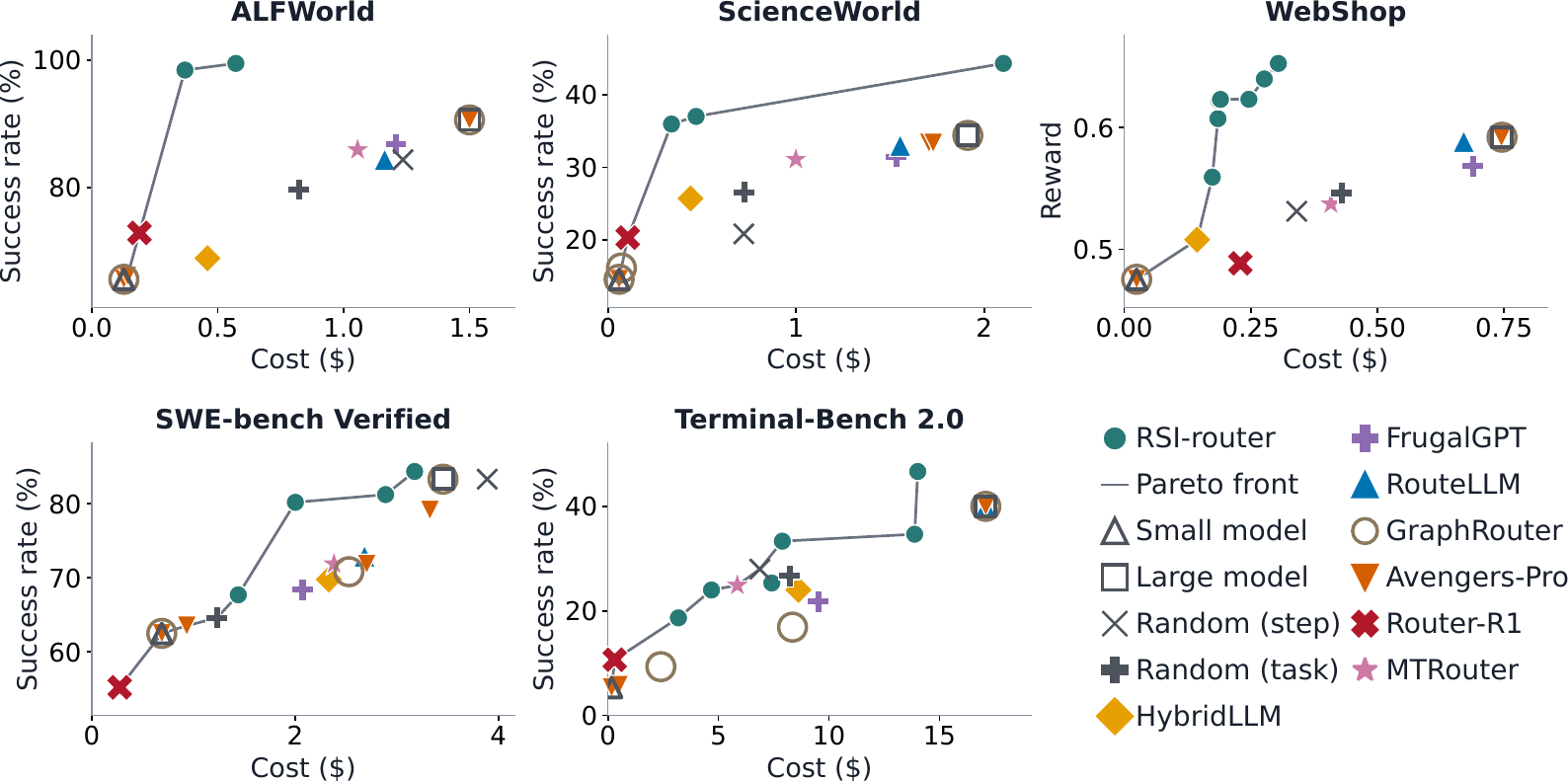}
\caption{\textbf{RSI-router occupies a substantial portion of the performance--cost Pareto frontier across five benchmarks.} Points show test performance and total inference cost (USD).}
\label{fig:9b-pareto}
\end{figure}

\textbf{RSI-router occupies a substantial portion of the performance--cost Pareto frontier across five benchmarks.}
While Table~1 compares the inference cost of different methods under a common performance constraint, Figure~\ref{fig:9b-pareto} provides a more comprehensive view of the performance--cost trade-off by plotting all evaluated candidate strategies for each method.
RSI-router improves the performance--cost Pareto frontier, with several configurations matching or outperforming the large model at lower cost.
By contrast, the step-level baselines Router-R1 and MTRouter reduce cost relative to the large model but remain below its performance on all five benchmarks under the evaluated settings (Appendix~\ref{app:baseline-implementation}).

\begin{figure}[htbp]
\centering
\includegraphics[width=\linewidth]{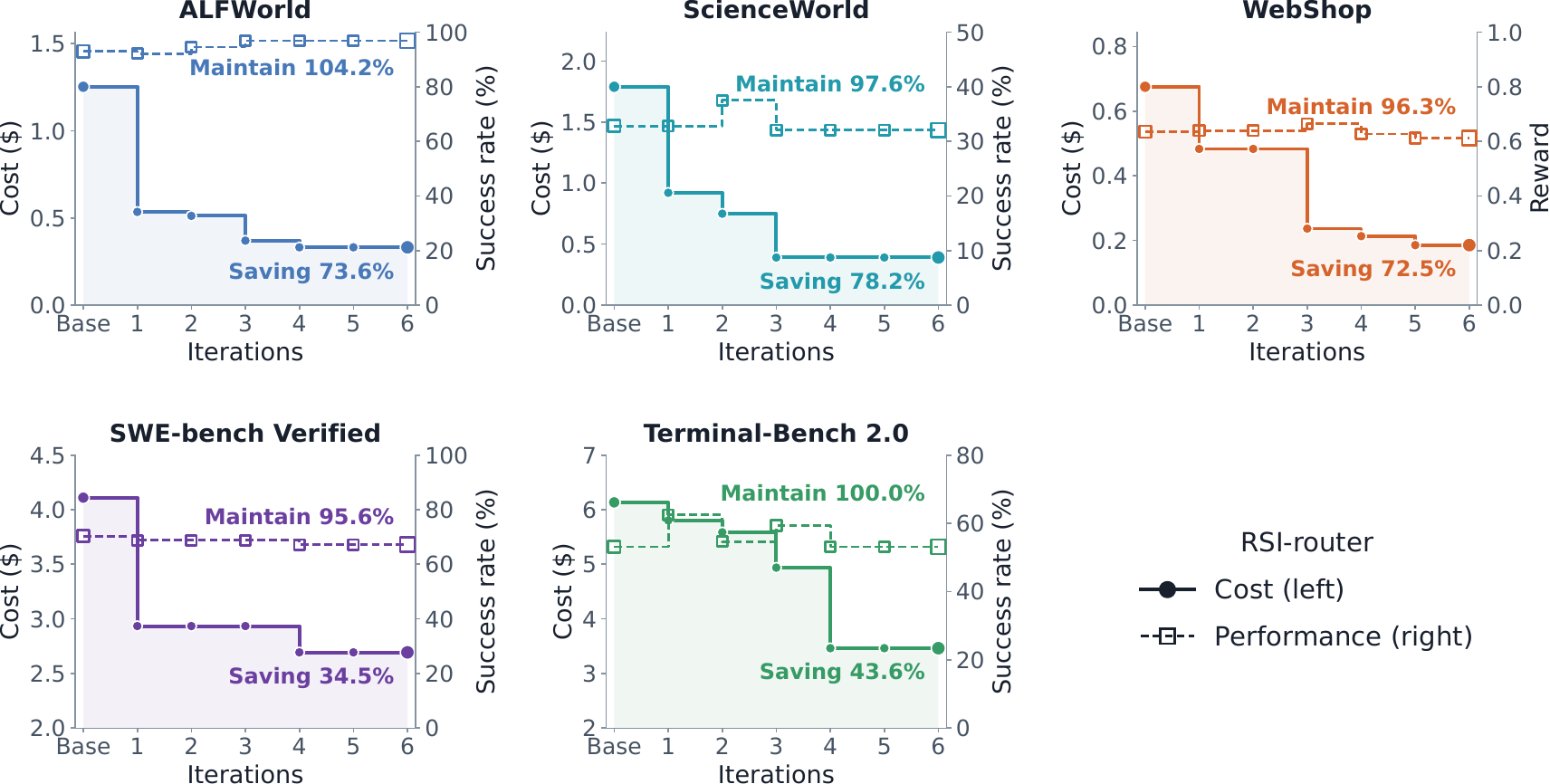}
\caption{\textbf{RSI-router progressively reduces cost during router evolution.} Solid curves show the lowest validation cost found through each iteration under a 95\% performance-retention constraint; dashed curves show the corresponding performance. Saving and Maintain report final cost reduction and performance retention relative to DeepSeek-V4.1-Flash.}
\label{fig:9b-cost-evolution}
\end{figure}

\begin{figure}[htbp]
\centering
\includegraphics[width=\linewidth]{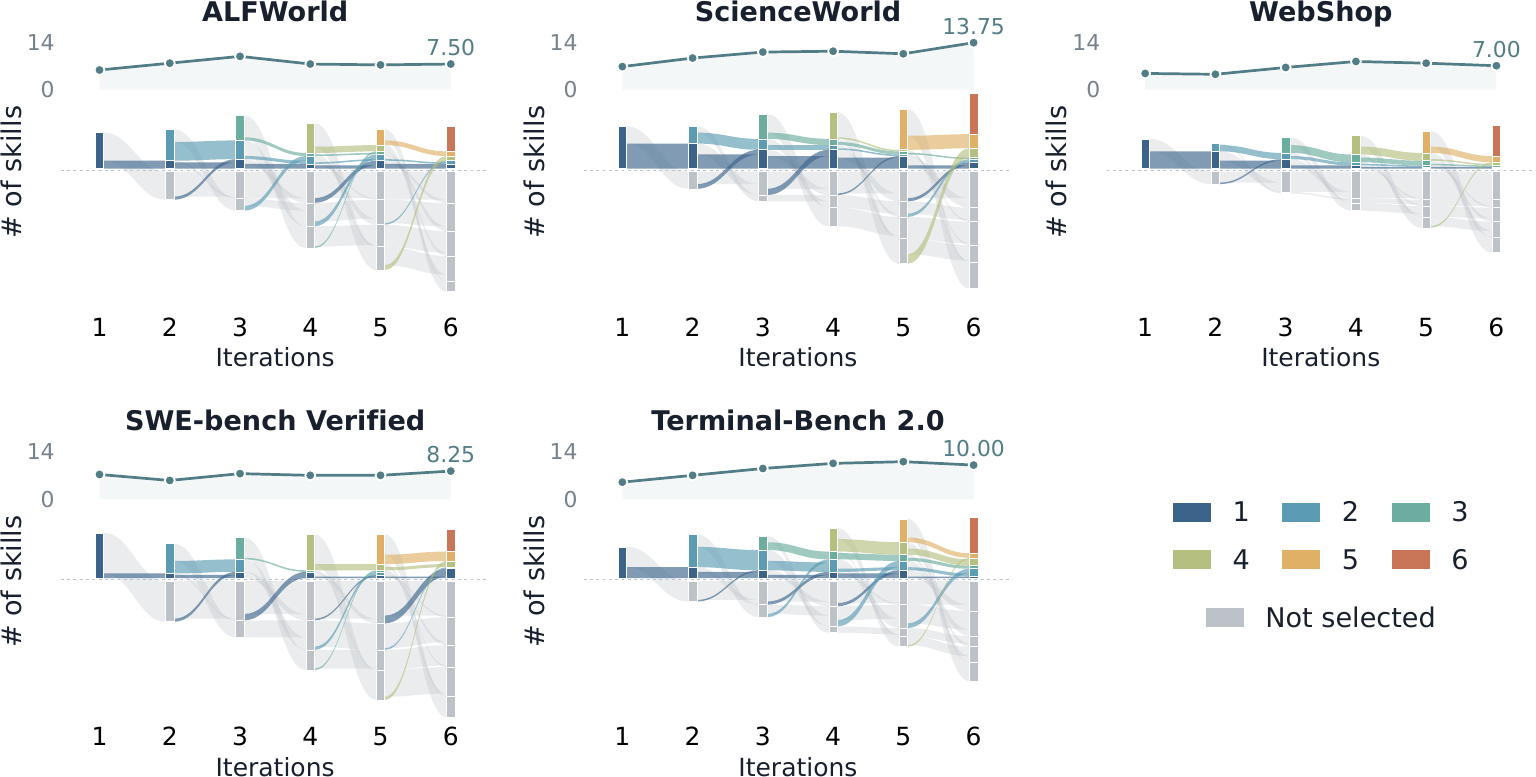}
\caption{\textbf{RSI-router accumulates new skills while reusing experience across iterations.} Upper curves show the mean number of execution skills per candidate router. Lower flows track unique skill sources across iterations.}
\label{fig:9b-skill-inheritance}
\end{figure}

\textbf{Several task-level baselines exhibit routing collapse and abrupt transitions.} In the evaluated settings, some baselines assign all tasks to either the small or the large model. As performance--cost preferences vary, these baselines switch abruptly between small-model-only and large-model-only routing, with no intermediate trade-offs observed. This pattern is consistent with whole-task selection overlooking the small model's ability to handle individual subtasks within tasks it cannot complete alone. In contrast, RSI-router enables the small model to contribute to tasks it cannot complete independently, better exploiting complementary model capabilities.



\textbf{RSI-router achieves repeated cost reductions across iterations.}
Figure~\ref{fig:9b-cost-evolution} tracks the lowest validation cost found through each iteration among routers retaining at least 95\% of the large-model baseline's performance. All five benchmarks show further cost reductions after the first iteration, with multiple reductions on four benchmarks. On WebShop, for example, cost savings increase from 28.5\% after the first iteration to 65.0\%, 68.4\%, and 72.5\% in subsequent iterations. These successive improvements demonstrate that continued router evolution discovers increasingly cost-efficient routers under the same performance constraint (see Appendix~\ref{app:router-evolution-cases}
for representative case studies).

\textbf{RSI-router accumulates reusable execution skills across iterations.}
Figure~\ref{fig:9b-skill-inheritance} tracks execution skills across the four candidate routers produced by Model-Specific Skill Evolution in each iteration. As new skills are added, skills acquired in earlier iterations continue to appear in later candidates. On ScienceWorld, for example, 22 of the 48 unique skill sources selected across the final iteration's candidates originate from earlier iterations. These results show that router evolution builds a skill library whose accumulated experience is reused in subsequent router construction.

\FloatBarrier
\subsection{Ablation Studies}
\label{sec:ablation-studies}

We conduct ablation studies to assess whether additional interaction history helps subtask identification and whether prompting the small model to select the execution model directly improves performance over step-level random routing.
We also remove execution skills from routers that contain them to assess their contribution to performance--cost trade-offs.

\begin{figure}[h]
\centering
\includegraphics[width=\linewidth]{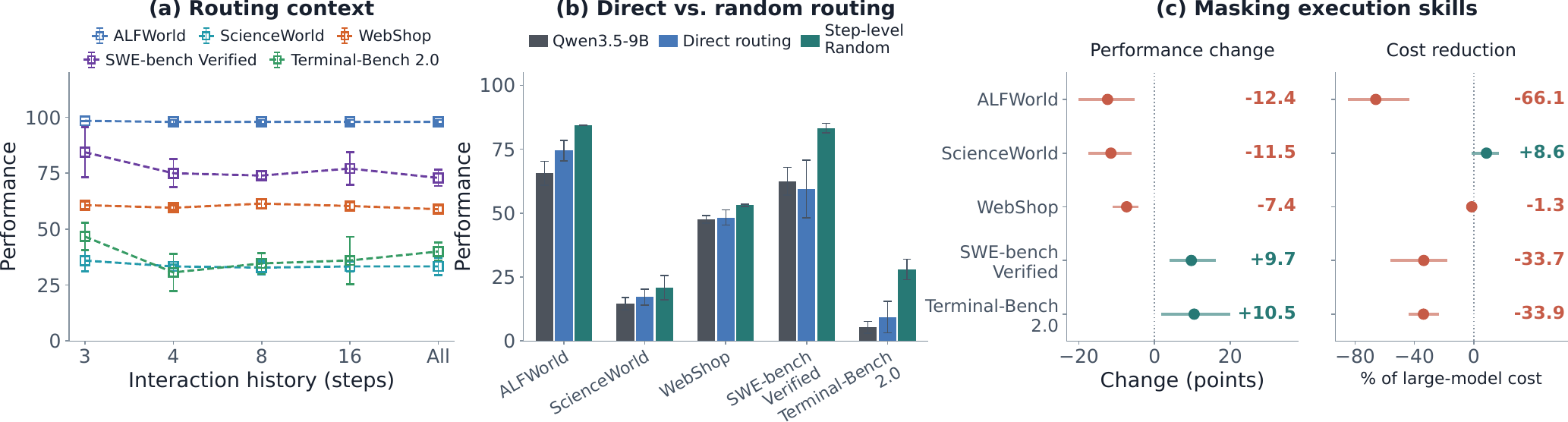}
\caption{\textbf{Ablations of routing context, direct routing, and execution skills.}
(a) Interaction history length.
(b) Direct model routing by Qwen3.5-9B.
(c) Effects of masking execution skills; cost reductions are normalized by the large-model-only USD cost.
WebShop reward is scaled by 100.
Error bars show standard deviations in (a--b) and 95\% bootstrap intervals in (c).}
\label{fig:9b-ablations}
\end{figure}

\textbf{Longer interaction history does not improve performance.}
For the routers reported in Table~\ref{tab:rsi-9b-grouped}, we increase the interaction history from 3 steps to 4, 8, 16, or all previous steps. As shown in Figure~\ref{fig:9b-ablations}(a), performance does not improve with longer interaction histories across the five benchmarks. This suggests that short histories are sufficient for effective routing.

\textbf{Direct difficulty-based routing underperforms step-level random routing.}
We provide Qwen3.5-9B with the same routing context used by RSI-router, but ask it to directly judge the difficulty of the upcoming step and select the execution model.
We compare it with the step-level Random baseline reported in Table~\ref{tab:rsi-9b-grouped}.
As shown in Figure~\ref{fig:9b-ablations}(b), direct routing achieves lower mean performance across all five benchmarks.
Thus, direct difficulty judgments do not yield a performance advantage over random step-level selection.

\textbf{Execution skills affect performance--cost trade-offs in task-dependent ways.}
To isolate their effect, we remove execution skills while keeping the underlying subtask definitions and model assignments unchanged. As shown in Figure~\ref{fig:9b-ablations}(c), removing skills lowers performance on ALFWorld, ScienceWorld, and WebShop by 12.4, 11.5, and 7.4 points, respectively.
On SWE-bench Verified and Terminal-Bench 2.0, it instead improves performance by 9.7 and 10.5 points, but increases cost by 33.7\% and 33.9\% of the respective large-model-only costs. These results indicate that execution skills do not uniformly improve performance; they shift the performance--cost trade-off by modifying how the assigned models execute each subtask.

\FloatBarrier

\section{Related Work}
\label{sec:related-work}

\paragraph{LLM routing.}
LLM routing exploits complementary model capabilities to improve performance--cost trade-offs~\citep{hu2024routerbench,li2026llmrouterbench,fernandez2026radar}.
At the task level, HybridLLM and RouteLLM select between small and large models using difficulty estimates and response preferences,
respectively~\citep{ding2024hybrid,ong2025routellm,song2025irt}.
FrugalGPT instead adopts cascaded inference, sequentially calling models until a response is judged sufficiently reliable~\citep{chen2024frugalgpt}. ICL-Router and IrtNet learn model representations to predict query-specific performance~\citep{wang2026icl,chen2025learning}. 
GraphRouter and Avengers use graph-based matching and cluster-level performance--cost estimates to select models~\citep{feng2025graphrouter,zhang2026avengers,zhang2025beyond}. 

Recent works extend routing to multi-step reasoning and agentic tasks. MTRouter learns model selection from sampled trajectories~\citep{zhang2026mtrouter}, while Router-R1 and Budget-Aware Agentic Routing optimize routing policies through reinforcement learning~\citep{zhang2025routerr,zhang2026budget}. SWE-Router uses partial execution traces to decide whether to switch to a more expensive model~\citep{son2026swe}. Agent-as-a-Router and Gated-Memory Routing inform model selection through accumulated outcomes and selective history retention, respectively~\citep{zhou2026agent,rajib2026learning}. SkillOrchestra extracts skills from pre-collected trajectories to guide the router's model selection~\citep{wang2026skillorchestra}. RSI-router uses subtasks to align model assignments with capability requirements and reuse experience across tasks. It leverages LLM prior knowledge to explore assignments and distills training trajectories into execution skills for the selected models.

\paragraph{Self-evolving agents.}
Agent self-improvement spans prompt and context adaptation, automated agent design, and harness evolution.
GEPA and ACE use execution feedback to refine prompts and contextual guidance, respectively~\citep{agrawal2026gepa,zhang2026agentic}.
ADAS~\citep{hu2025automated} searches over agent programs, while the Darwin G\"odel Machine~\citep{zhang2026darwin} iteratively modifies agent code and maintains an expanding archive of candidates. At the harness level, Meta-Harness~\citep{lee2026meta} uses an external meta-agent powered by a strong model to optimize the target model's harness. Self-Harness~\citep{zhang2026self} enables agents to autonomously diagnose their weaknesses and evolve their own harnesses, without relying on human engineers or stronger external agents.
Harness-of-Harness~\citep{yan2026harness} organizes coding-agent executions into iterative planning, coding, and testing loops to support continual software improvement.
HarnessBank~\citep{luo2607harnessbank} maintains a semantically diverse harness population, combining candidate recombination with gated verification.
HarnessX~\citep{chen2026harnessx} evolves composable harness components from execution traces, while Harness Continual Learning~\citep{kang2026harness} supports continual harness adaptation under historical-retention constraints.
We adopt this feedback-driven evolution process for agentic routing itself, recursively refining model selection and execution guidance from execution outcomes and accumulated experience.

\paragraph{Agent skills.}
Agent skills provide reusable procedural knowledge for task execution, and their benefits vary across models~\citep{li2026skillsbench}.
Prior work distills skills from execution experience~\citep{ni2026trace2skill,xia2026skillrl}.
AgentSkillOS and SkillNet organize skills for retrieval and orchestration~\citep{li2026organizing,liang2026skillnet}, while other work trains agents to internalize and apply skills~\citep{lu2026skill0,zhu2026skill0}.
SKT further generates verified synthetic trajectories for supervised skill-use training~\citep{tan2026skt}.
RSI-router evolves model-specific skills alongside subtask-level routing strategies to help smaller models execute their assigned subtasks more reliably.

\section{Conclusion}
\label{sec:conclusion}

We introduced \textit{RSI-router}, a routing framework that combines LLM task knowledge with execution feedback to improve performance--cost trade-offs for agentic tasks.
It organizes model assignments around reusable subtasks and uses comparisons with large-model-only trajectories to develop model-specific execution skills.
Candidate updates are assessed through re-execution and Pareto-based router selection using validation performance and cost, while historical routers and skills guide subsequent iterations.

Across five agentic benchmarks, RSI-router reduces inference cost by 8.1--82.2\% while matching or exceeding large-model-only performance and extends the performance--cost Pareto frontier beyond the evaluated task-level and step-level routing baselines.
These results demonstrate that smaller models can contribute to tasks they cannot complete independently when assigned suitable subtasks and supported by execution skills.
Further analysis shows repeated cost reductions and continued reuse of historical skills, highlighting the value of accumulated execution experience.

More broadly, these findings support using iterative evolution
to construct routing strategies for agentic tasks. They highlight the potential to improve agentic performance and efficiency by better exploiting complementary model capabilities, allowing smaller models to contribute to tasks beyond what they can complete independently. RSI-router provides a foundation for constructing cost-aware routing strategies from accumulated execution experience, reducing reliance on continual manual refinement for agentic tasks.

\section*{Acknowledgements}
This work is supported by Shanghai Artificial Intelligence Laboratory.

\bibliographystyle{unsrtnat}
\bibliography{references}

@inproceedings{
shridhar2021alfworld,
title={{\{}ALFW{\}}orld: Aligning Text and Embodied Environments for Interactive Learning},
author={Mohit Shridhar and Xingdi Yuan and Marc-Alexandre Cote and Yonatan Bisk and Adam Trischler and Matthew Hausknecht},
booktitle={International Conference on Learning Representations},
year={2021},
url={https://openreview.net/forum?id=0IOX0YcCdTn}
}

@inproceedings{wang2022scienceworld,
  title={Scienceworld: Is your agent smarter than a 5th grader?},
  author={Wang, Ruoyao and Jansen, Peter and C{\^o}t{\'e}, Marc-Alexandre and Ammanabrolu, Prithviraj},
  booktitle={Proceedings of the 2022 Conference on Empirical Methods in Natural Language Processing},
  pages={11279--11298},
  year={2022}
}

@article{yao2022webshop,
  title={Webshop: Towards scalable real-world web interaction with grounded language agents},
  author={Yao, Shunyu and Chen, Howard and Yang, John and Narasimhan, Karthik},
  journal={Advances in Neural Information Processing Systems},
  volume={35},
  pages={20744--20757},
  year={2022}
}

@inproceedings{jimenez2024swe,
  title={Swe-bench: Can language models resolve real-world github issues?},
  author={Jimenez, Carlos E and Yang, John and Wettig, Alexander and Yao, Shunyu and Pei, Kexin and Press, Ofir and Narasimhan, Karthik},
  booktitle={International Conference on Learning Representations},
  volume={2024},
  pages={54107--54157},
  year={2024}
}

@inproceedings{merrill2026terminal,
  title={Terminal-bench: Benchmarking agents on hard, realistic tasks in command line interfaces},
  author={Merrill, Mike and Shaw, Alexander and Carlini, Nicholas and Li, Boxuan and Raj, Harsh and Bercovich, Ivan and Shi, Lin and Shin, Jeong and Walshe, Thomas and Buchanan, E Kelly and others},
  booktitle={International Conference on Learning Representations},
  volume={2026},
  pages={40903--40986},
  year={2026}
}

@inproceedings{ding2024hybrid,
  title={Hybrid llm: Cost-efficient and quality-aware query routing},
  author={Ding, Dujian and Mallick, Ankur and Wang, Chi and Sim, Robert and Mukherjee, Subhabrata and R{\"u}hle, Victor and Lakshmanan, Laks and Awadallah, Ahmed H},
  booktitle={International Conference on Learning Representations},
  volume={2024},
  pages={41348--41366},
  year={2024}
}

@article{
chen2024frugalgpt,
title={Frugal{GPT}: How to Use Large Language Models While Reducing Cost and Improving Performance},
author={Lingjiao Chen and Matei Zaharia and James Zou},
journal={Transactions on Machine Learning Research},
issn={2835-8856},
year={2024},
url={https://openreview.net/forum?id=cSimKw5p6R}
}

@inproceedings{ong2025routellm,
  title={Routellm: Learning to route llms from preference data},
  author={Ong, Isaac and Almahairi, Amjad and Wu, Vincent and Chiang, Wei-Lin and Wu, Tianhao and Gonzalez, Joseph E and Kadous, Mohammed and Stoica, Ion},
  booktitle={International Conference on Learning Representations},
  volume={2025},
  pages={34433--34448},
  year={2025}
}

@inproceedings{feng2025graphrouter,
  title={Graphrouter: A graph-based router for llm selections},
  author={Feng, Tao and Shen, Yanzhen and You, Jiaxuan},
  booktitle={International Conference on Learning Representations},
  volume={2025},
  pages={26186--26203},
  year={2025}
}

@inproceedings{zhang2025beyond,
  title={Beyond gpt-5: Making llms cheaper and better via performance-efficiency optimized routing},
  author={Zhang, Yiqun and Li, Hao and Chen, Jianhao and Zhang, Hangfan and Ye, Peng and Bai, Lei and Hu, Shuyue},
  booktitle={Proceedings of the 2025 7th International Conference on Distributed Artificial Intelligence},
  pages={122--129},
  year={2025}
}

@inproceedings{zhang2026mtrouter,
  title={MTRouter: Cost-Aware Multi-Turn LLM Routing with History--Model Joint Embeddings},
  author={Zhang, Yiqun and Li, Hao and Wang, Zihan and Feng, Shi and Yang, Xiaocui and Wang, Daling and Zhang, Bo and Bai, Lei and Hu, Shuyue},
  booktitle={Proceedings of the 64th Annual Meeting of the Association for Computational Linguistics (Volume 1: Long Papers)},
  pages={44206--44226},
  year={2026}
}

@article{yang2024swe,
  title={Swe-agent: Agent-computer interfaces enable automated software engineering},
  author={Yang, John and Jimenez, Carlos and Wettig, Alexander and Lieret, Kilian and Yao, Shunyu and Narasimhan, Karthik and Press, Ofir},
  journal={Advances in Neural Information Processing Systems},
  volume={37},
  pages={50528--50652},
  year={2024}
}

@misc{deepseek2026v41flash,
  author       = {{DeepSeek-AI}},
  title        = {{DeepSeek-V4.1-Flash}: Pushing the Limits of {KV} Cache Compression},
  year         = {2026},
  howpublished = {Hugging Face model card},
  url          = {https://huggingface.co/deepseek-ai/DeepSeek-V4.1-Flash},
  note         = {Accessed: 2026-09-17}
}

@misc{qwen2026qwen35,
  author       = {{Qwen Team}},
  title        = {{Qwen3.5}: Towards Native Multimodal Agents},
  year         = {2026},
  month        = feb,
  url          = {https://qwen.ai/blog?id=qwen3.5},
  note         = {Accessed: 2026-09-17}
}

@misc{openai2025codex,
  author       = {{OpenAI}},
  title        = {Introducing {Codex}},
  year         = {2025},
  month        = may,
  url          = {https://openai.com/index/introducing-codex/},
  note         = {Accessed: 2026-09-17}
}

@misc{chowdhury2024swebenchverified,
  author       = {Chowdhury, Neil and Aung, James and Shern, Chan Jun
                  and Jaffe, Oliver and Sherburn, Dane and Starace, Giulio
                  and Mays, Evan and Dias, Rachel and Aljubeh, Marwan
                  and Glaese, Mia and Jimenez, Carlos E. and Yang, John
                  and Ho, Leyton and Patwardhan, Tejal and Liu, Kevin
                  and Madry, Aleksander},
  title        = {Introducing {SWE-bench Verified}},
  year         = {2024},
  month        = aug,
  howpublished = {OpenAI},
  url          = {https://openai.com/index/introducing-swe-bench-verified/},
  note         = {Accessed: 2026-09-17}
}

@article{hu2024routerbench,
  title={Routerbench: A benchmark for multi-llm routing system},
  author={Hu, Qitian Jason and Bieker, Jacob and Li, Xiuyu and Jiang, Nan and Keigwin, Benjamin and Ranganath, Gaurav and Keutzer, Kurt and Upadhyay, Shriyash Kaustubh},
  journal={arXiv preprint arXiv:2403.12031},
  year={2024}
}

@inproceedings{li2026llmrouterbench,
  title={LLMRouterBench: A massive benchmark and unified framework for LLM routing},
  author={Li, Hao and Zhang, Yiqun and Guo, Zhaoyan and Wang, Chenxu and Tang, Shengji and Zhang, Qiaosheng and Chen, Yang and Qi, Biqing and Ye, Peng and Bai, Lei and others},
  booktitle={Findings of the Association for Computational Linguistics: ACL 2026},
  pages={37733--37754},
  year={2026}
}

@article{zhang2026budget,
  title={Budget-aware agentic routing via boundary-guided training},
  author={Zhang, Caiqi and Xia, Menglin and Zhang, Xuchao and Madrigal, Daniel and Mallick, Ankur and Kessler, Samuel and Ruehle, Victor and Rajmohan, Saravan},
  journal={arXiv preprint arXiv:2602.21227},
  year={2026}
}

@article{son2026swe,
  title={SWE-Router: Routing in Multi-turn Agentic Software Engineering Tasks},
  author={Son, Seongho and Yoon, Sangwoong and Tang, Jiahua and Wang, Shuhan and Wolf, Lorenz and Bogunovic, Ilija},
  journal={arXiv preprint arXiv:2607.00053},
  year={2026}
}

@article{zhou2026agent,
  title={Agent-as-a-Router: Agentic Model Routing for Coding Tasks},
  author={Zhou, Pengfei and Tang, Zhiwei and Ma, Yixing and Tang, Jiasheng and Han, Yizeng and Wan, Zhenglin and Meng, Fanqing and Wang, Wei and Zhuang, Bohan and Zhao, Wangbo and others},
  journal={arXiv preprint arXiv:2606.22902},
  year={2026}
}

@article{rajib2026learning,
  title={Learning What to Retain: Gated-Memory Routing for Efficient Collaboration in Multi-Agent LLM Systems},
  author={Rajib, Rakibul Hasan and Zheng, Mengxing and Lou, Qian},
  journal={arXiv preprint arXiv:2609.00237},
  year={2026}
}

@article{wang2026skillorchestra,
  title={Skillorchestra: Learning to route agents via skill transfer},
  author={Wang, Jiayu and Ming, Yifei and Ke, Zixuan and Joty, Shafiq and Albarghouthi, Aws and Sala, Frederic},
  journal={arXiv preprint arXiv:2602.19672},
  year={2026}
}

@inproceedings{agrawal2026gepa,
  title={Gepa: Reflective prompt evolution can outperform reinforcement learning},
  author={Agrawal, Lakshya A and Tan, Shangyin and Soylu, Dilara and Ziems, Noah and Khare, Rishi and Opsahl-Ong, Krista and Singhvi, Arnav and Shandilya, Herumb and Ryan, Michael J and Jiang, Meng and others},
  booktitle={International Conference on Learning Representations},
  volume={2026},
  pages={8479--8565},
  year={2026}
}

@inproceedings{zhang2026agentic,
  title={Agentic context engineering: Evolving contexts for self-improving language models},
  author={Zhang, Qizheng and Hu, Changran and Upasani, Shubhangi and Ma, Boyuan and Hong, Fenglu and Kamanuru, Vamsidhar and Rainton, Jay and Wu, Chen and Ji, Mengmeng and Li, Hanchen and others},
  booktitle={International Conference on Learning Representations},
  volume={2026},
  pages={86069--86100},
  year={2026}
}

@inproceedings{hu2025automated,
  title={Automated design of agentic systems},
  author={Hu, Shengran and Lu, Cong and Clune, Jeff},
  booktitle={International Conference on Learning Representations},
  volume={2025},
  pages={21344--21377},
  year={2025}
}

@inproceedings{
zhang2026darwin,
title={Darwin G\"odel Machine: Open-Ended Evolution of Self-Improving Agents},
author={Jenny Zhang and Shengran Hu and Cong Lu and Robert Tjarko Lange and Jeff Clune},
booktitle={The Fourteenth International Conference on Learning Representations},
year={2026},
url={https://openreview.net/forum?id=pUpzQZTvGY}
}

@article{lee2026meta,
  title={Meta-harness: End-to-end optimization of model harnesses},
  author={Lee, Yoonho and Nair, Roshen and Zhang, Qizheng and Lee, Kangwook and Khattab, Omar and Finn, Chelsea},
  journal={arXiv preprint arXiv:2603.28052},
  year={2026}
}

@article{zhang2026self,
  title={Self-harness: Harnesses that improve themselves},
  author={Zhang, Hangfan and Zhang, Shao and Li, Kangcong and Zhang, Chen and Chen, Yang and Zhang, Yiqun and Bai, Lei and Hu, Shuyue},
  journal={arXiv preprint arXiv:2606.09498},
  year={2026}
}

@article{luo2607harnessbank,
  title={Harnessbank: Semantic gene-bank search with gated verification for agent-harness self-evolution, 2026},
  author={Luo, Xiaotian and Xue, Dizhan and Wang, Fengxingyu and Hu, Chuanrui and Deng, Yafeng},
  journal={URL https://arxiv. org/abs/2607.13683}
}

@article{chen2026harnessx,
  title={Harnessx: A composable, adaptive, and evolvable agent harness foundry},
  author={Chen, Tingyang and Lu, Shuo and Zhao, Kang and Meng, Weicheng and Teng, Hanlin and Li, Tianhao and Li, Chao and Liu, Xule and Liang, Jian and Zhang, Zhizhong and others},
  journal={arXiv preprint arXiv:2606.14249},
  year={2026}
}

@article{kang2026harness,
  title={Harness Continual Learning: Continual Adaptation Beyond Model Parameters},
  author={Kang, Borui and Gu, Jinrui and Lv, Junhan and Li, Wenbin and Wang, Lei and Gao, Yang},
  journal={arXiv preprint arXiv:2608.19013},
  year={2026}
}

@inproceedings{
zhang2025routerr,
title={Router-R1: Teaching {LLM}s Multi-Round Routing and Aggregation via Reinforcement Learning},
author={Haozhen Zhang and Tao Feng and Jiaxuan You},
booktitle={The Thirty-ninth Annual Conference on Neural Information Processing Systems},
year={2025},
url={https://openreview.net/forum?id=DWf4vroKWJ}
}

@article{liu2024large,
  title={Large language model-based agents for software engineering: A survey},
  author={Liu, Junwei and Wang, Kaixin and Chen, Yixuan and Peng, Xin and Chen, Zhenpeng and Zhang, Lingming and Lou, Yiling},
  journal={ACM Transactions on Software Engineering and Methodology},
  year={2024},
  publisher={ACM New York, NY}
}

@article{sager2026comprehensive,
  title={A comprehensive survey of agents for computer use: Foundations, challenges, and future directions},
  author={Sager, Pascal J and Meyer, Benjamin and Yan, Peng and von Wartburg-Kottler, Rebekka and Etaiwi, Layan and Enayati, Aref and Nobel, Gabriel and Abdulkadir, Ahmed and Grewe, Benjamin F and Stadelmann, Thilo},
  journal={Journal of Artificial Intelligence Research},
  volume={85},
  year={2026}
}

@article{lu2026towards,
  title={Towards end-to-end automation of AI research},
  author={Lu, Chris and Lu, Cong and Lange, Robert Tjarko and Yamada, Yutaro and Hu, Shengran and Foerster, Jakob and Ha, David and Clune, Jeff},
  journal={Nature},
  volume={651},
  number={8107},
  pages={914--919},
  year={2026},
  publisher={Nature Publishing Group UK London}
}

@inproceedings{jiang2023llm,
  title={Llm-blender: Ensembling large language models with pairwise ranking and generative fusion},
  author={Jiang, Dongfu and Ren, Xiang and Lin, Bill Yuchen},
  booktitle={Proceedings of the 61st Annual Meeting of the Association for Computational Linguistics (Volume 1: Long Papers)},
  pages={14165--14178},
  year={2023}
}

@inproceedings{jitkrittum2026universal,
  title={Universal model routing for efficient llm inference},
  author={Jitkrittum, Wittawat and Narasimhan, Harikrishna and Rawat, Ankit Singh and Juneja, Jeevesh and Wang, Congchao and Wang, Zifeng and Go, Alec and Lee, Chen-Yu and Shenoy, Pradeep and Panigrahy, Rina and others},
  booktitle={International Conference on Learning Representations},
  volume={2026},
  pages={10169--10218},
  year={2026}
}

@inproceedings{gao2026more,
  title={More with less: An empirical study of turn-control strategies for efficient coding agents},
  author={Gao, Pengfei and Peng, Chao},
  booktitle={Proceedings of the 2026 IEEE/ACM 48th International Conference on Software Engineering},
  pages={956--967},
  year={2026}
}

@inproceedings{
alazraki2026scaling,
title={Scaling Small Agents Through Strategy Auctions},
author={Lisa Alazraki and William F. Shen and Yoram Bachrach and Akhil Mathur},
booktitle={Forty-third International Conference on Machine Learning},
year={2026},
url={https://openreview.net/forum?id=elXuA5wTWV}
}

@article{abouelenin2025phi,
  title={Phi-4-mini technical report: Compact yet powerful multimodal language models via mixture-of-loras},
  author={Abouelenin, Abdelrahman and Ashfaq, Atabak and Atkinson, Adam and Awadalla, Hany and Bach, Nguyen and Bao, Jianmin and Benhaim, Alon and Cai, Martin and Chaudhary, Vishrav and Chen, Congcong and others},
  journal={arXiv preprint arXiv:2503.01743},
  year={2025}
}

@inproceedings{song2025irt,
  title={Irt-router: Effective and interpretable multi-llm routing via item response theory},
  author={Song, Wei and Huang, Zhenya and Cheng, Cheng and Gao, Weibo and Xu, Bihan and Zhao, GuanHao and Wang, Fei and Wu, Runze},
  booktitle={Proceedings of the 63rd Annual Meeting of the Association for Computational Linguistics (Volume 1: Long Papers)},
  pages={15629--15644},
  year={2025}
}

@inproceedings{fernandez2026radar,
  title={Radar: Reasoning-ability and difficulty-aware routing for reasoning llms},
  author={Fernandez, Nigel Steven and Kveton, Branislav and Rossi, Ryan and Lan, Andrew and Wang, Jack},
  booktitle={International Conference on Learning Representations},
  volume={2026},
  pages={109429--109457},
  year={2026}
}

@article{zhang2024advancing,
  title={Advancing DRL agents in commercial fighting games: Training, integration, and agent-human alignment},
  author={Zhang, Chen and He, Qiang and Yuan, Zhou and Liu, Elvis S and Wang, Hong and Zhao, Jian and Wang, Yang},
  journal={arXiv preprint arXiv:2406.01103},
  year={2024}
}

@article{tan2026skt,
  title={SKT: Skill-use training at scale via verified synthetic data generation},
  author={Tan, Zelin and Zhang, Yiqun and Li, Hao and Cui, Zhiyao and Geng, Hejia and Zhang, Shao and Zhang, Hangfan and Chen, Yang and Wang, Xiaosong and Wang, Lilong and others},
  journal={arXiv preprint arXiv:2608.02287},
  year={2026}
}

@inproceedings{tan2026scaling,
  title={Scaling behaviors of llm reinforcement learning post-training: An empirical study in mathematical reasoning},
  author={Tan, Zelin and Geng, Hejia and Yu, Xiaohang and Zhang, Mulei and Wan, Guancheng and Zhou, Yifan and He, Qiang and Xue, Xiangyuan and Zhou, Heng and Fan, Yutao and others},
  booktitle={Proceedings of the 64th Annual Meeting of the Association for Computational Linguistics (Volume 1: Long Papers)},
  pages={31300--31319},
  year={2026}
}

@article{tan2026papo,
  title={PAPO: Stabilizing Rubric Integration Training via Decoupled Advantage Normalization},
  author={Tan, Zelin and Yu, Zhouliang and Lin, Bohan and Geng, Zijie and Geng, Hejia and Zhang, Yudong and Zhang, Mulei and Chen, Yang and Hu, Shuyue and Yin, Zhenfei and others},
  journal={arXiv preprint arXiv:2603.26535},
  year={2026}
}

@article{cui2026agentpanel,
  title={AgentPanel: Toward a New Paradigm for Human--AI Collaboration in Exploring Scientific Questions},
  author={Cui, Zhiyao and Wang, Qianyi and Yan, Haoyang and Zhang, Yiqun and Ren, Siyue and Zhang, Hangfan and Tan, Zelin and Li, Hao and Mu, Chunjiang and Cai, Dexian and others},
  journal={arXiv preprint arXiv:2608.03283},
  year={2026}
}

@inproceedings{wang2026icl,
  title={Icl-router: In-context learned model representations for llm routing},
  author={Wang, Chenxu and Li, Hao and Zhang, Yiqun and Chen, Linyao and Chen, Jianhao and Jian, Ping and Zhang, Qiaosheng and Hu, Shuyue},
  booktitle={Proceedings of the AAAI Conference on Artificial Intelligence},
  volume={40},
  number={39},
  pages={33413--33421},
  year={2026}
}

@article{chen2025learning,
  title={Learning compact representations of LLM abilities via item response theory},
  author={Chen, Jianhao and Wang, Chenxu and Zhang, Gengrui and Ye, Peng and Bai, Lei and Hu, Wei and Qu, Yuzhong and Hu, Shuyue},
  journal={arXiv preprint arXiv:2510.00844},
  year={2025}
}

@article{li2026organizing,
  title={Organizing, orchestrating, and benchmarking agent skills at ecosystem scale},
  author={Li, Hao and Mu, Chunjiang and Chen, Jianhao and Ren, Siyue and Cui, Zhiyao and Zhang, Yiqun and Bai, Lei and Hu, Shuyue},
  journal={arXiv preprint arXiv:2603.02176},
  year={2026}
}

@inproceedings{zhang2026avengers,
  title={The Avengers: A Routing Recipe for Collective Intelligence in Language Models},
  author={Zhang, Yiqun and Li, Hao and Wang, Chenxu and Chen, Linyao and Zhang, Qiaosheng and Ye, Peng and Feng, Shi and Wang, Xinrun and Xu, Jia and Bai, Lei and others},
  booktitle={Proceedings of the AAAI Conference on Artificial Intelligence},
  volume={40},
  number={41},
  pages={34870--34878},
  year={2026}
}

@article{yan2026harness,
  title={Harness-of-Harness: Multi-Day Autonomous Software Development with Continual Improvement},
  author={Yan, Haoyang and Su, Min-le and Zhang, Hangfan and Li, Zhanhao and Zhang, Chen and Zhang, Shao and Chen, Yang and Bai, Lei and Hu, Shuyue},
  journal={arXiv preprint arXiv:2609.01481},
  year={2026}
}

@inproceedings{cui2026design,
  title={Design First, Code Later: Aesthetically Pleasing Template-Free Slides Generation},
  author={Cui, Zhiyao and Wang, Chenxu and Hu, Shuyue and Zhang, Yiqun and Shao, Wenqi and Zhang, Qiaosheng and Wang, Zhen},
  booktitle={Findings of the Association for Computational Linguistics: ACL 2026},
  pages={30470--30490},
  year={2026}
}

@article{li2026skillsbench,
  title={SkillsBench: Benchmarking how well agent skills work across diverse tasks},
  author={Li, Xiangyi and Liu, Yimin and Chen, Wenbo and You, Bingran and Di, Zonglin and He, Yifeng and Zheng, Shenghan and Choe, Kyoung Whan and Sun, Jiankai and Wang, Shuyi and others},
  journal={arXiv preprint arXiv:2602.12670},
  year={2026}
}

@article{ni2026trace2skill,
  title={Trace2skill: Distill trajectory-local lessons into transferable agent skills},
  author={Ni, Jingwei and Liu, Yihao and Liu, Xinpeng and Sun, Yutao and Zhou, Mengyu and Cheng, Pengyu and Wang, Dexin and Zhao, Erchao and Jiang, Xiaoxi and Jiang, Guanjun},
  journal={arXiv preprint arXiv:2603.25158},
  year={2026}
}

@article{xia2026skillrl,
  title={Skillrl: Evolving agents via recursive skill-augmented reinforcement learning},
  author={Xia, Peng and Chen, Jianwen and Wang, Hanyang and Liu, Jiaqi and Zeng, Kaide and Wang, Yu and Han, Siwei and Zhou, Yiyang and Zhao, Xujiang and Chen, Haifeng and others},
  journal={arXiv preprint arXiv:2602.08234},
  year={2026}
}

@article{lu2026skill0,
  title={Skill0: In-context agentic reinforcement learning for skill internalization},
  author={Lu, Zhengxi and Yao, Zhiyuan and Wu, Jinyang and Han, Chengcheng and Gu, Qi and Cai, Xunliang and Lu, Weiming and Xiao, Jun and Zhuang, Yueting and Shen, Yongliang},
  journal={arXiv preprint arXiv:2604.02268},
  year={2026}
}

@article{liang2026skillnet,
  title={Skillnet: Create, evaluate, and connect ai skills},
  author={Liang, Yuan and Zhong, Ruobin and Xu, Haoming and Jiang, Chen and Zhong, Yi and Fang, Runnan and Gu, Jia-Chen and Deng, Shumin and Yao, Yunzhi and Wang, Mengru and others},
  journal={arXiv preprint arXiv:2603.04448},
  year={2026}
}

@article{zhu2026skill0,
  title={Skill0. 5: Joint Skill Internalization and Utilization for Out-of-Distribution Generalization in Agentic Reinforcement Learning},
  author={Zhu, Jiapeng and Yu, Jianxiang and Zhao, Yibo and Han, Chengcheng and Gu, Qi and Cai, Xunliang and Li, Xiang and Qian, Weining},
  journal={arXiv preprint arXiv:2605.28424},
  year={2026}
}

\clearpage
\appendix
\begin{center}
  {\LARGE\sffamily\bfseries Appendix}
\end{center}
\section{Baseline Implementation Details}
\label{app:baseline-implementation}

\paragraph{Shared setup.}
We use the official implementations of all task-level and step-level routing baselines.
All methods route between $L$ and $S$ under the experimental setup in Section~\ref{sec:experimental-setup}.
For each benchmark, we fit the task-level and step-level routing baselines on $\mathcal D_{\mathrm{tr}}\cup\mathcal D_{\mathrm{val}}$ and evaluate all methods on $\mathcal D_{\mathrm{test}}$. All results are averaged over three evaluation repeats per test task. For learned baselines trained with three seeds, we additionally average across these training seeds.

\paragraph{Fixed policies.}
The large-model-only router $r_L$ and the small-model-only router $r_S$ use $L$ and $S$, respectively, throughout task execution.
The task-level random router selects either model with equal probability once per task and uses it throughout execution.
The step-level random router selects either model with equal probability before each execution step.
These baselines require no router training.

\paragraph{HybridLLM.}
We use HybridLLM~\citep{ding2024hybrid} with a Qwen3-Embedding-0.6B encoder and a binary classification head to estimate task difficulty.
The encoder and classification head are jointly fine-tuned using AdamW, with a learning rate of $10^{-5}$, weight decay of $0.1$, and batch size 4.
Gradient accumulation gives an effective batch size of 64.
The warmup ratio is 0.05, and the tolerated performance gap between the two models is set to zero.

\paragraph{FrugalGPT.}
We adapt FrugalGPT~\citep{chen2024frugalgpt} to a task-level cascade.
For each execution model, we fine-tune a separate Qwen3-Embedding-0.6B scorer that predicts task performance from the user query, using mean observed task performance as soft supervision.
Each scorer is trained using AdamW, with a learning rate of $2\times10^{-5}$, weight decay of $0.01$, batch size 4, and a warmup ratio of 0.06.
Cost includes the task execution costs of all models invoked by the cascade.

\paragraph{RouteLLM.}
We use the matrix-factorization router of RouteLLM~\citep{ong2025routellm}, with frozen Qwen3-Embedding-8B query embeddings projected into a 128-dimensional space.
Model embeddings and the projection are trained using pairwise preferences derived from observed task performance.
Optimization uses Adam with a learning rate of $3\times10^{-4}$, weight decay of $10^{-5}$, and batch size 64.
At inference time, the estimated preference probability determines model selection.

\paragraph{GraphRouter.}
We use GraphRouter~\citep{feng2025graphrouter} for graph-based model selection, with Qwen3-Embedding-8B representations of queries, tasks, and model descriptions.
All models tied for the highest training utility receive positive labels.
Training uses AdamW, with a learning rate of $3\times10^{-4}$ and weight decay of $10^{-4}$.
We evaluate the three settings defined in the original paper: \emph{Performance First}, \emph{Balance}, and \emph{Cost First}.

\paragraph{Avengers-Pro.}
We use Avengers-Pro~\citep{zhang2025beyond} with Qwen3-Embedding-8B query embeddings and $k$-means clustering with $k=8$.
Models are ranked using cluster-level performance--cost estimates, and each query is routed according to the ranking in its nearest cluster.
We evaluate ten evenly spaced performance coefficients in $\{0,1/9,\ldots,1\}$, with the complementary weight assigned to cost.

\paragraph{Router-R1.}
We adapt Router-R1~\citep{zhang2025routerr} with Qwen3.5-9B as the policy model.
At each step, the policy interleaves reasoning with model calls and integrates the returned responses to generate the next action.
Training uses PPO with task-level outcome rewards, actor and critic learning rates of $10^{-6}$ and $10^{-5}$, respectively, and a KL penalty coefficient of $10^{-3}$.
We use a rollout batch size of 64 and a PPO mini-batch size of 32.

\paragraph{MTRouter.}
We train MTRouter~\citep{zhang2026mtrouter} on trajectories collected by $r_L$, $r_S$, and the step-level random router.
It selects a model at each step using the interaction history, encoded by frozen Qwen3-Embedding-8B. The outcome network has two hidden layers of width 256 and dropout 0.1.
Training uses AdamW with early stopping, a learning rate of $10^{-3}$, weight decay of $0.01$, and batch size 64.

\section{Benchmark Prompts}
\label{app:benchmark-prompts}

\begingroup
\setlength{\parindent}{0pt}
\setlength{\parskip}{5pt}
\raggedbottom

\Needspace{8\baselineskip}
\subsection{ALFWorld}
\label{app:prompts-alfworld}

\begin{benchmarkprompt}{ALFWorld System Prompt}
You are a helpful assistant interacting with ALFWorld, a text-based household environment. Your goal is to complete the household task by issuing text commands.\par
\bpGap
\bpHeading{How to Interact}
Write exactly one command in a \bpInlineCode{\bpBacktick{}\bpBacktick{}\bpBacktick{}text\bpBacktick{}\bpBacktick{}\bpBacktick{}} code block. The environment will respond with observations.\par
\bpGap
\begin{bpLiteral}
<format\_example>\par
THOUGHT: I should inspect the room and choose a valid action.\par
\vspace{3pt}
\bpBacktick{}\bpBacktick{}\bpBacktick{}text\par
look\par
\bpBacktick{}\bpBacktick{}\bpBacktick{}\par
</format\_example>\par
\end{bpLiteral}
\bpGap
\bpHeading{Important Rules}
\begin{bpList}
\item Issue ONE command per response.
\item Prefer commands from the available actions list shown in the observation.
\item If no available actions are shown, issue the most likely ALFWorld command.
\item Do not claim the task is complete yourself; keep acting until the environment ends the episode.
\end{bpList}
\end{benchmarkprompt}

\begin{benchmarkprompt}{ALFWorld User Prompt}
\bpTemplateLine{<task\_description>\\*\{\{task\_description\}\}\\*</task\_description>}
\bpGap
\bpTemplateLine{<task\_type>\\*\{\{task\_type\}\}\\*</task\_type>}
\bpGap
\bpTemplateLine{<initial\_observation>\\*\{\{initial\_observation\}\}\\*</initial\_observation>}
\bpGap
\begin{bpLiteral}
\{\% if valid\_actions \%\}\par
<available\_actions>\par
\{\% for action in valid\_actions \%\}\par
- \{\{ action \}\}\par
\{\% endfor \%\}\par
</available\_actions>\par
\{\% endif \%\}\par
\end{bpLiteral}
\bpGap
Complete the ALFWorld task. Think briefly, then choose exactly one command.\par
\end{benchmarkprompt}

\Needspace{8\baselineskip}
\subsection{ScienceWorld}
\label{app:prompts-scienceworld}

\begin{benchmarkprompt}{ScienceWorld System Prompt}
You are a helpful assistant interacting with a text-based science simulation environment. Your goal is to complete science experiments by issuing text commands.\par
\bpGap
\bpHeading{How to Interact}
You issue commands by writing them in a \bpInlineCode{\bpBacktick{}\bpBacktick{}\bpBacktick{}text\bpBacktick{}\bpBacktick{}\bpBacktick{}} code block. The environment will respond with observations.\par
\bpGap
\begin{bpLiteral}
<format\_example>\par
THOUGHT: I should explore my surroundings first.\par
\vspace{3pt}
\bpBacktick{}\bpBacktick{}\bpBacktick{}text\par
look around\par
\bpBacktick{}\bpBacktick{}\bpBacktick{}\par
</format\_example>\par
\end{bpLiteral}
\bpGap
\bpHeading{Available Command Types}
Common commands include:\par
\begin{bpList}
\item Movement: \bpInlineCode{"go to [location]"}, \bpInlineCode{"open door"}, \bpInlineCode{"go through door"}
\item Interaction: \bpInlineCode{"pick up [object]"}, \bpInlineCode{"put [object] in [container]"}, \bpInlineCode{"activate [object]"}
\item Observation: \bpInlineCode{"look around"}, \bpInlineCode{"look at [object]"}, \bpInlineCode{"inventory"}
\item Task-specific: \bpInlineCode{"focus on [object]"}, \bpInlineCode{"use [tool] on [object]"}
\end{bpList}
\bpGap
\bpHeading{Query Commands (Free Actions)}
You can query available actions without consuming a game turn:\par
\begin{bpList}
\item \bpInlineCode{?navigation} - Show movement actions (go, walk, move)
\item \bpInlineCode{?object} - Show object manipulation (pick up, put, pour)
\item \bpInlineCode{?observation} - Show observation actions (look, examine, inventory)
\item \bpInlineCode{?device} - Show device control (activate, turn on/off, use)
\item \bpInlineCode{?door} - Show door/container actions (open, close)
\item \bpInlineCode{?electrical} - Show electrical actions (connect, disconnect)
\item \bpInlineCode{?interaction} - Show interaction actions (mix, eat, focus)
\item \bpInlineCode{?all} - Show all valid actions
\item \bpInlineCode{?categories} - Show query help
\end{bpList}
\bpGap
Use queries to explore available actions before deciding your next move.\par
\bpGap
\bpHeading{Important Notes}
\begin{bpList}
\item Issue ONE command per response
\item Include a THOUGHT section explaining your reasoning
\item The environment is turn-based - wait for observations before issuing the next command
\item Some tasks require multiple steps to complete
\end{bpList}
\end{benchmarkprompt}

\begin{benchmarkprompt}{ScienceWorld User Prompt}
\bpTemplateLine{<task\_description>\\*\{\{task\_description\}\}\\*</task\_description>}
\bpGap
\bpTemplateLine{<initial\_observation>\\*\{\{initial\_observation\}\}\\*</initial\_observation>}
\bpGap
\bpTemplateLine{<instructions>}\nopagebreak[4]
Complete the science experiment described above. You are interacting with a simulated environment. Issue commands one at a time and observe the results.\par
\bpGap
\textbf{Tip:} Use query commands like \bpInlineCode{?navigation} or \bpInlineCode{?object} to explore available actions without consuming a turn.\par
\bpGap
\bpHeading{\textbf{Strategy hints:}}
\begin{bpList}
\item First explore: use \bpInlineCode{"open door to [room]"} then \bpInlineCode{"go to [room]"} to navigate
\item Look around each room to find objects you need
\item Pick up objects with \bpInlineCode{"pick up [object]"}
\item For heating: find a stove, turn it on with \bpInlineCode{"activate [stove]"}, place container on it
\item For cooling: use a freezer or refrigerator
\item Use \bpInlineCode{"focus on [object]"} to examine substances
\end{bpList}
\bpGap
To complete the task, perform the necessary actions described in the task description. When you believe the task is complete, issue the command:\par
\bpGap
\begin{bpLiteral}
\bpBacktick{}\bpBacktick{}\bpBacktick{}text\par
task completed\par
\bpBacktick{}\bpBacktick{}\bpBacktick{}\par
\end{bpLiteral}
\bpGap
Remember:\par
\begin{bpList}
\item Think step by step about what actions are needed
\item Explore your environment to find needed objects
\item Some actions may require prerequisites (e.g., picking up an object before using it)
\end{bpList}
\bpTemplateLine{</instructions>}
\end{benchmarkprompt}

\Needspace{8\baselineskip}
\subsection{WebShop}
\label{app:prompts-webshop}

\begin{benchmarkprompt}{WebShop System Prompt}
You are an expert autonomous agent operating in the WebShop e-commerce environment. Your goal is to buy the product that best satisfies the user instruction.\par
\bpGap
Valid command formats are:\par
\begin{bpList}
\item search[keywords]
\item click[button or product id or option]
\end{bpList}
\bpGap
For each step, think internally if needed, then output exactly one admissible WebShop command in \bpInlineCode{<action>}...\bpInlineCode{</action>}.\par
\bpGap
The command inside \bpInlineCode{<action>} must be chosen from the current admissible actions list, except \bpInlineCode{search[...]} may replace \bpInlineCode{search[<your query>]} with useful keywords.\par
\bpGap
Your final visible response must contain the \bpInlineCode{<action>}...\bpInlineCode{</action>} tag. Do not spend the whole response on hidden reasoning.\par
Do not claim the task is complete yourself; keep acting until the environment ends the episode.\par
\end{benchmarkprompt}

\begin{benchmarkprompt}{WebShop User Prompt}
\bpTemplateLine{<task\_description>\\*\{\{task\_description\}\}\\*</task\_description>}
\bpGap
\bpTemplateLine{<task\_type>\\*\{\{task\_type\}\}\\*</task\_type>}
\bpGap
\bpTemplateLine{<history>\\*\{\{action\_history\}\}\\*</history>}
\bpGap
\bpTemplateLine{<current\_observation>\\*\{\{current\_observation\}\}\\*</current\_observation>}
\bpGap
\bpTemplateLine{<admissible\_actions>\\*\{\{valid\_actions\_text\}\}\\*</admissible\_actions>}
\bpGap
You are at WebShop step \bpInlineCode{\{\{current\_step\}\}}. Choose one action now.\par
\bpGap
\begin{bpLiteral}
<format\_example>\par
<think>I should search for the requested product type.</think>\par
<action>search[black running shoes]</action>\par
</format\_example>\par
\end{bpLiteral}
\end{benchmarkprompt}

\Needspace{8\baselineskip}
\subsection{SWE-bench Verified}
\label{app:prompts-swebench}

The assistant issues native \bpInlineCode{bash} tool calls with a \bpInlineCode{command} argument.
Execution results are returned as \bpInlineCode{tool} messages.

\begin{benchmarkprompt}{SWE-bench Verified System Prompt}
You are a helpful assistant that can interact with a computer shell to solve programming tasks.\par
\end{benchmarkprompt}

\begin{benchmarkprompt}{SWE-bench Verified User Prompt}
\bpTemplateLine{<pr\_description>}\nopagebreak[4]
Consider the following PR description:\par
\bpTemplateLine{\{\{task\}\}}
\bpTemplateLine{</pr\_description>}
\bpGap
\bpTemplateLine{<instructions>}\nopagebreak[4]
\bpHeading{Task Instructions}
\bpGap
\bpHeading{Overview}
\bpGap
You're a software engineer interacting continuously with a computer by submitting commands. You'll be helping implement necessary changes to meet requirements in the PR description. Your task is specifically to make changes to non-test files in the current directory in order to fix the issue described in the PR description in a way that is general and consistent with the codebase.\par
\bpInlineCode{<IMPORTANT>}This is an interactive process where you will think and issue AT LEAST ONE command, see the result, then think and issue your next command(s).\bpInlineCode{</important>}\par
\bpGap
For each response:\par
\bpGap
\begin{bpList}
\item[1.] Include a THOUGHT section explaining your reasoning and what you're trying to accomplish
\item[2.] Provide one or more bash tool calls to execute
\end{bpList}
\bpGap
\bpHeading{Important Boundaries}
\bpGap
\begin{bpList}
\item MODIFY: Regular source code files in /testbed (this is the working directory for all your subsequent commands)
\item DO NOT MODIFY: Tests, configuration files (pyproject.toml, setup.cfg, etc.)
\end{bpList}
\bpGap
\bpHeading{Recommended Workflow}
\bpGap
\begin{bpList}
\item[1.] Analyze the codebase by finding and reading relevant files
\item[2.] Create a script to reproduce the issue
\item[3.] Edit the source code to resolve the issue
\item[4.] Verify your fix works by running your script again
\item[5.] Test edge cases to ensure your fix is robust
\end{bpList}
\bpGap
\bpHeading{Command Execution Rules}
\bpGap
You are operating in an environment where\par
\bpGap
\begin{bpList}
\item[1.] You issue at least one command
\item[2.] The system executes the command(s) in a subshell
\item[3.] You see the result(s)
\item[4.] You write your next command(s)
\end{bpList}
\bpGap
Each response should include:\par
\bpGap
\begin{bpList}
\item[1.] \textbf{Reasoning text} where you explain your analysis and plan
\item[2.] At least one tool call with your command
\end{bpList}
\bpGap
\bpHeading{\textbf{CRITICAL REQUIREMENTS:}}
\bpGap
\begin{bpList}
\item Your response SHOULD include reasoning text explaining what you're doing
\item Your response MUST include AT LEAST ONE bash tool call. You can make MULTIPLE tool calls in a single response when the commands are independent (e.g., searching multiple files, reading different parts of the codebase).
\item Directory or environment variable changes are not persistent. Every action is executed in a new subshell.
\item However, you can prefix any action with \bpInlineCode{MY\_ENV\_VAR=MY\_VALUE cd /path/to/working/dir \&\& ...} or write/load environment variables from files
\end{bpList}
\bpGap
Example of a CORRECT response:\par
\begin{bpLiteral}
<example\_response>\par
I need to understand the Builder-related code. Let me find relevant files and check the project structure.\par
\vspace{3pt}
[Makes multiple bash tool calls: \{"command": "ls -la"\}, \{"command": "find src -name \bpSingleQuote{}*.java\bpSingleQuote{} | grep -i builder"\}, \{"command": "cat README.md | head -50"\}]\par
</example\_response>\par
\end{bpLiteral}
\bpGap
\bpHeading{Environment Details}
\bpGap
\begin{bpList}
\item You have a full Linux shell environment
\item Always use non-interactive flags (-y, -f) for commands
\item Avoid interactive tools like vi, nano, or any that require user input
\item You can use bash commands or invoke any tool that is available in the environment
\item You can also create new tools or scripts to help you with the task
\item If a tool isn't available, you can also install it
\end{bpList}
\bpGap
\bpHeading{Submission}
\bpGap
When you've completed your work, you MUST submit your changes as a git patch. Follow these steps IN ORDER, with SEPARATE commands:\par
\bpGap
\bpHeading{Step 1: Create the patch file}
Run \bpInlineCode{git diff -{}- path/to/file1 path/to/file2 > patch.txt} listing only the source files you modified. Do NOT commit your changes.\par
\bpGap
\bpTemplateLine{<IMPORTANT>}\nopagebreak[4]
The patch must only contain changes to the specific source files you modified to fix the issue. Do not submit file creations or changes to any of the following files:\par
\bpGap
\begin{bpList}
\item test and reproduction files
\item helper scripts, tests, or tools that you created
\item installation, build, packaging, configuration, or setup scripts unless they are directly part of the issue you were fixing (you can assume that the environment is already set up for your client)
\item binary or compiled files
\end{bpList}
\bpTemplateLine{</IMPORTANT>}
\bpGap
\bpHeading{Step 2: Verify your patch}
Inspect patch.txt to confirm it only contains your intended changes and headers show \bpInlineCode{-{}-{}- a/} and \bpInlineCode{+++ b/} paths.\par
\bpGap
\bpHeading{Step 3: Submit (EXACT command required)}
You MUST use this EXACT command to submit:\par
\bpGap
\begin{bpLiteral}
\bpBacktick{}\bpBacktick{}\bpBacktick{}bash\par
echo COMPLETE\_TASK\_AND\_SUBMIT\_FINAL\_OUTPUT \&\& cat patch.txt\par
\bpBacktick{}\bpBacktick{}\bpBacktick{}\par
\end{bpLiteral}
\bpGap
If the command fails (nonzero exit status), it will not submit.\par
\bpGap
\bpTemplateLine{<CRITICAL>}\nopagebreak[4]
\begin{bpList}
\item Creating/viewing the patch and submitting it MUST be separate commands (not combined with \&\&).
\item If you modify patch.txt after verifying, you SHOULD verify again before submitting.
\item You CANNOT continue working (reading, editing, testing) in any way on this task after submitting.
\end{bpList}
\bpTemplateLine{</CRITICAL>}
\bpTemplateLine{</instructions>}
\end{benchmarkprompt}

\begin{benchmarktool}{SWE-bench Verified}
\begin{Verbatim}[fontsize=\small,breaklines=true,breakanywhere=true,breaksymbolleft={},breaksymbolright={}]
{
  "type": "function",
  "function": {
    "name": "bash",
    "description": "Execute a bash command",
    "parameters": {
      "type": "object",
      "properties": {
        "command": {
          "type": "string",
          "description": "The bash command to execute"
        }
      },
      "required": [
        "command"
      ]
    }
  }
}
\end{Verbatim}
\end{benchmarktool}

\Needspace{8\baselineskip}
\subsection{Terminal-Bench 2.0}
\label{app:prompts-tb2}

The assistant issues native \bpInlineCode{bash} tool calls with a \bpInlineCode{command} argument.
Execution results are returned as \bpInlineCode{tool} messages.

\begin{benchmarkprompt}{Terminal-Bench 2.0 System Prompt}
You are a helpful assistant that solves tasks by interacting with a computer shell.\par
\end{benchmarkprompt}

\begin{benchmarkprompt}{Terminal-Bench 2.0 User Prompt}
\bpTemplateLine{\{\{task\}\}}
\bpGap
Work in the current task environment and continue until the task is fully complete.\par
\bpGap
Each response must include reasoning about the next step and at least one bash tool call. Directory and environment-variable changes are not persistent between tool calls, so include them in each command when needed.\par
\bpGap
When the task is complete, run \bpInlineCode{echo COMPLETE\_TASK\_AND\_SUBMIT\_FINAL\_OUTPUT} as a standalone command. Do not combine it with another command; after it runs you cannot continue working.\par
\bpGap
\bpTemplateLine{<system\_information>\\*\{\{system\}\} \{\{release\}\} \{\{version\}\} \{\{machine\}\}\\*</system\_information>}
\end{benchmarkprompt}

\begin{benchmarktool}{Terminal-Bench 2.0}
\begin{Verbatim}[fontsize=\small,breaklines=true,breakanywhere=true,breaksymbolleft={},breaksymbolright={}]
{
  "type": "function",
  "function": {
    "name": "bash",
    "description": "Execute a bash command",
    "parameters": {
      "type": "object",
      "properties": {
        "command": {
          "type": "string",
          "description": "The bash command to execute"
        }
      },
      "required": [
        "command"
      ]
    }
  }
}
\end{Verbatim}
\end{benchmarktool}

\par
\endgroup

\section{Estimating Small-Model Cost from GPU Hours}
\label{app:gpu-hour-cost}

\paragraph{Pricing reference.}
We use the official DeepSeek-V4.1-Flash API prices as the USD reference: \$0.30 per million input tokens and \$1.20 per million output tokens.\footnote{\url{https://api-docs.deepseek.com/quick_start/pricing/}}
We estimate Qwen3.5-9B's relative cost using GPU-hour measurements, then express its input and output prices relative to these official rates.

\paragraph{Measurement protocol.}
We benchmark DeepSeek-V4.1-Flash on eight NVIDIA H200 GPUs and Qwen3.5-9B on one H200 using SGLang.
Both deployments receive identical synthetic texts at three reference input lengths and two fixed output lengths (Table~\ref{tab:gpu-hour-calibration}).
Each setting uses 1,024 requests with 64 concurrent requests, replenished as requests finish.
For model $m$ using $n_m$ GPUs over $T_m$ seconds, we compute
\begin{equation}
G_m=\frac{n_m T_m}{3600}\quad\text{GPU hours}.
\end{equation}

\begin{table}[htbp]
\centering
\small
\caption{GPU hours for 1,024 matched requests per model at concurrency 64. Input lengths use the Qwen3.5-9B tokenizer as reference. Each setting is measured once. Ratios are DeepSeek/Qwen.}
\label{tab:gpu-hour-calibration}
\renewcommand{\arraystretch}{1.12}
\begin{tabular*}{\linewidth}{@{\extracolsep{\fill}}ccccc@{}}
\toprule
\multicolumn{2}{c}{Tokens per request} & \multicolumn{2}{c}{GPU hours} & \\
\cmidrule(lr){1-2}\cmidrule(lr){3-4}
Input & Output & DeepSeek-V4.1-Flash & Qwen3.5-9B & Ratio \\
      &        & 8 H200s             & 1 H200      &       \\
\midrule
2,048  & 128   & 0.9842  & 0.0199 & 49.46 \\
2,048  & 1,024 & 5.4078  & 0.0545 & 99.23 \\
8,192  & 128   & 2.0729  & 0.0679 & 30.53 \\
8,192  & 1,024 & 6.4912  & 0.1153 & 56.30 \\
32,768 & 128   & 7.1375  & 0.2819 & 25.32 \\
32,768 & 1,024 & 11.5576 & 0.3962 & 29.17 \\
\midrule
\multicolumn{2}{l}{All six settings} & 33.6512 & 0.9357 & 35.96 \\
\bottomrule
\end{tabular*}
\end{table}

\paragraph{Cost estimate.}
Across the six synthetic workloads, the total GPU-hour ratio between DeepSeek-V4.1-Flash and Qwen3.5-9B is 35.96.
We use $1{:}30$ as an approximate reference for pricing Qwen3.5-9B, yielding input and output rates of \$0.01 and \$0.04 per million tokens, respectively.

\section{Case Studies of Router Evolution}
\label{app:router-evolution-cases}
\begin{figure}[!htbp]
\centering
\includegraphics[width=\linewidth]{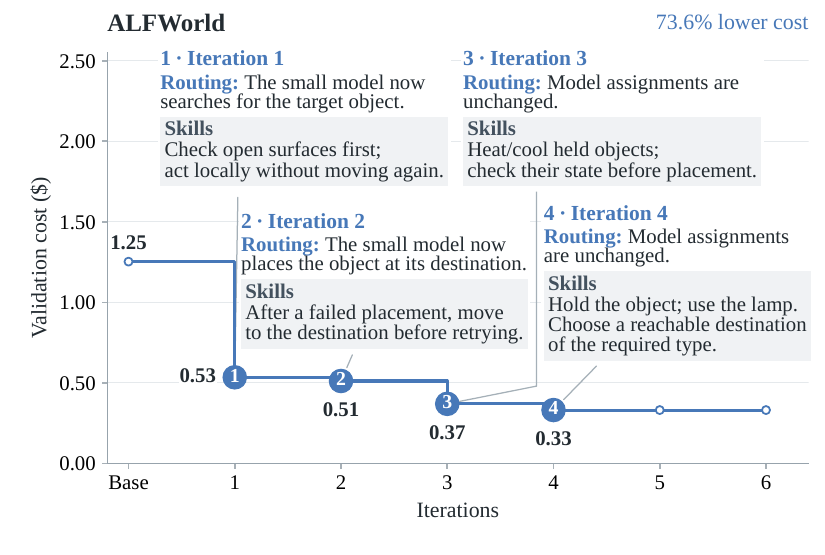}
\caption{\textbf{Router evolution on ALFWorld.}}
\label{fig:evolution-alfworld}
\vspace{-1em}
\end{figure}

\begin{figure}[!htbp]
\centering
\includegraphics[width=\linewidth]{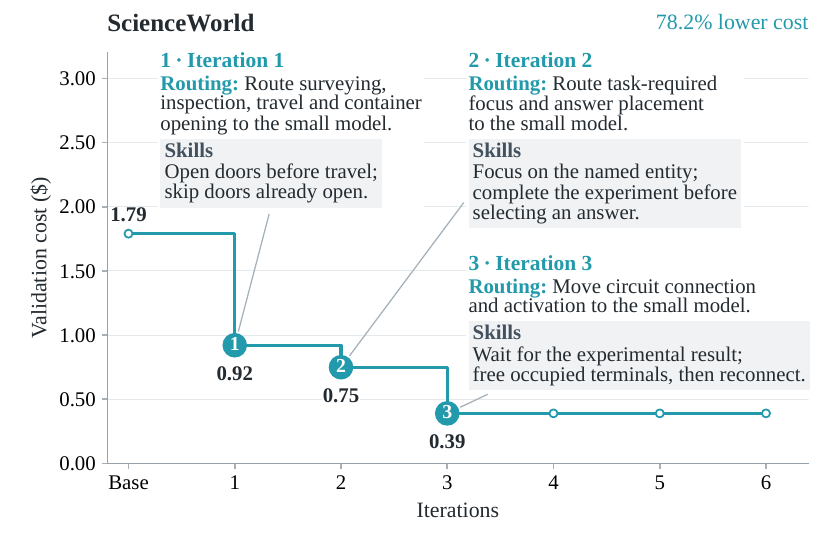}
\caption{\textbf{Router evolution on ScienceWorld.}}
\label{fig:evolution-scienceworld}
\vspace{-1em}
\end{figure}

\begin{figure}[!htbp]
\centering
\includegraphics[width=\linewidth]{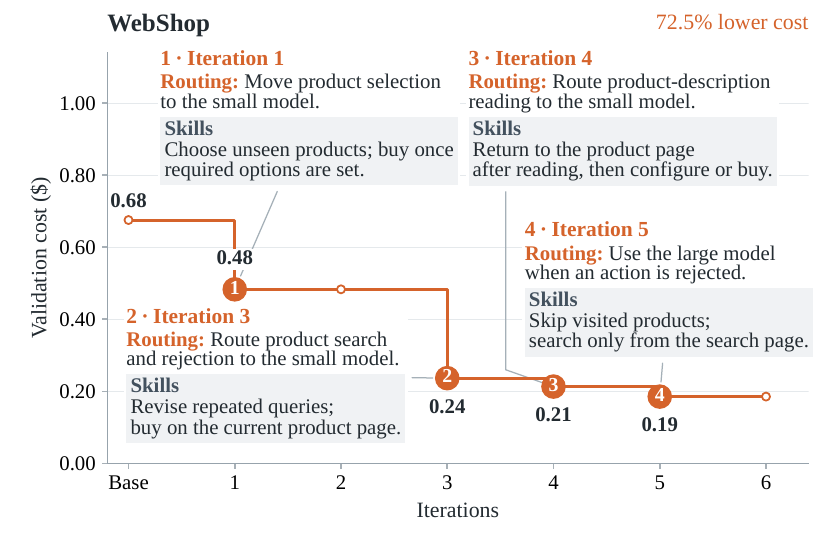}
\caption{\textbf{Router evolution on WebShop.}}
\label{fig:evolution-webshop}
\vspace{-1em}
\end{figure}

\begin{figure}[!htbp]
\centering
\includegraphics[width=\linewidth]{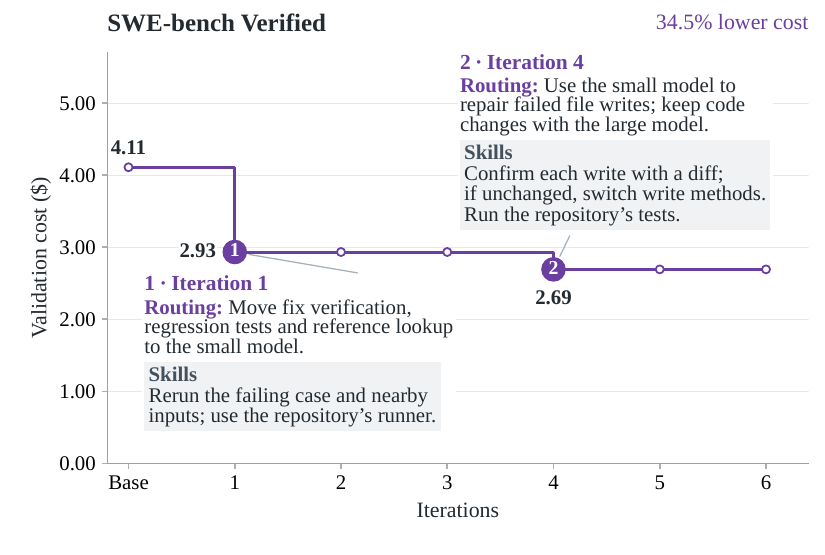}
\caption{\textbf{Router evolution on SWE-bench Verified.}}
\label{fig:evolution-swebench}
\vspace{-1em}
\end{figure}

\begin{figure}[!htbp]
\centering
\includegraphics[width=\linewidth]{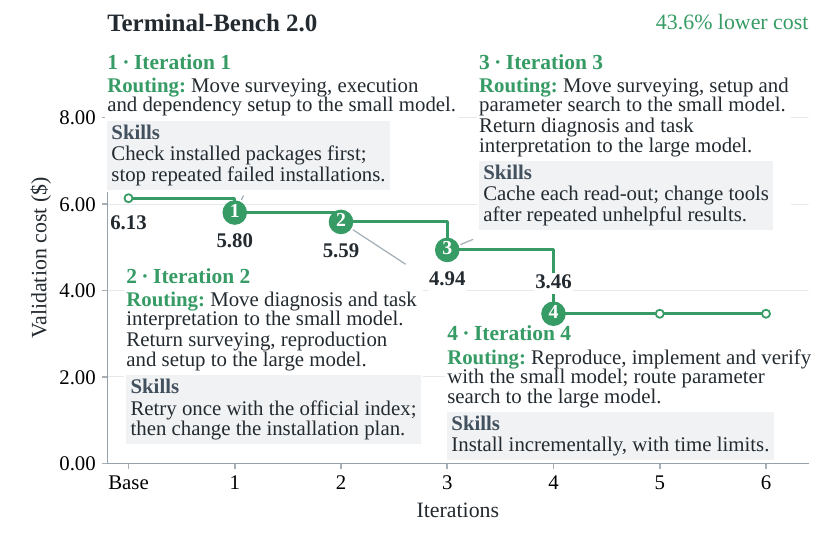}
\caption{\textbf{Router evolution on Terminal-Bench 2.0.}}
\label{fig:evolution-tb2}
\vspace{-1em}
\end{figure}

We illustrate representative routing changes and execution skills
during router evolution in Figures~\ref{fig:evolution-alfworld}--\ref{fig:evolution-tb2}.

\end{document}